\documentclass[10pt,twocolumn,letterpaper]{article}

\usepackage[pagenumbers]{iccv} % To force page numbers, e.g. for an arXiv version
\usepackage[accsupp]{axessibility}
\definecolor{iccvblue}{rgb}{0.21,0.49,0.74}
\usepackage[pagebackref,breaklinks,colorlinks,allcolors=iccvblue]{hyperref}
\usepackage{bm}
\usepackage{array}
\usepackage{multirow}
\usepackage{float}

\def\paperID{6877} % *** Enter the Paper ID here
\def\confName{ICCV}
\def\confYear{2025}

\title{Unsupervised Part Discovery via Descriptor-Based Masked Image Restoration with Optimized Constraints}

\author{Jiahao Xia$^{1}$, Yike Wu$^{1}$, Wenjian Huang$^{2}$, Jianguo Zhang$^{2}$, Jian Zhang\thanks{Corresponding Author} $^{ ,1}$\\
	{$^1$Faculty of Engineering and IT, University of Technology Sydney}\\
	{$^2$ Dept. of Comp. Sci. and Eng., Southern University of Science and Technology}\\
	{\small Jiahao.Xia-1@uts.edu.au, Yike.Wu@student.uts.edu.au, \{huangwj, zhangjg\}@sustech.edu.cn, Jian.Zhang@uts.edu.au}
	\\}

\begin{document}
\maketitle

\begin{abstract} 
	Part-level features are crucial for image understanding, but few studies focus on them because of the lack of fine-grained labels. Although unsupervised part discovery can eliminate the reliance on labels, most of them cannot maintain robustness across various categories and scenarios, which restricts their application range. To overcome this limitation, we present a more effective paradigm for unsupervised part discovery, named \textbf{M}asked \textbf{P}art \textbf{A}uto\textbf{e}ncoder (MPAE). It first learns part descriptors as well as a feature map from the inputs and produces patch features from a masked version of the original images. Then, the masked regions are filled with the learned part descriptors based on the similarity between the local features and descriptors. By restoring these masked patches using the part descriptors, they become better aligned with their part shapes, guided by appearance features from unmasked patches. Finally, MPAE robustly discovers meaningful parts that closely match the actual object shapes, even in complex scenarios. Moreover, several looser yet more effective constraints are proposed to enable MPAE to identify the presence of parts across various scenarios and categories in an unsupervised manner. This provides the foundation for addressing challenges posed by occlusion and for exploring part similarity across multiple categories. Extensive experiments demonstrate that our method robustly discovers meaningful parts across various categories and scenarios. The code is available at the project website\footnote{\url{https://github.com/Jiahao-UTS/MPAE}}.
\end{abstract}

\section{Introduction}

Decomposing objects into meaningful parts drives models to gain better image understanding and further improves the performance of downstream tasks, such as person search~\cite{PLOT}, fine-grained classifcation~\cite{PDiscoNet, PDiscoFormer} and behavior analysis \cite{Animal_Kindom}. But the high cost of part-level label results in a scarcity of training samples. Consequently, part-level feature learning has received much less attention compared to instance-level feature learning.

\begin{figure}[t!]
	\centering
	\includegraphics[width=\linewidth]{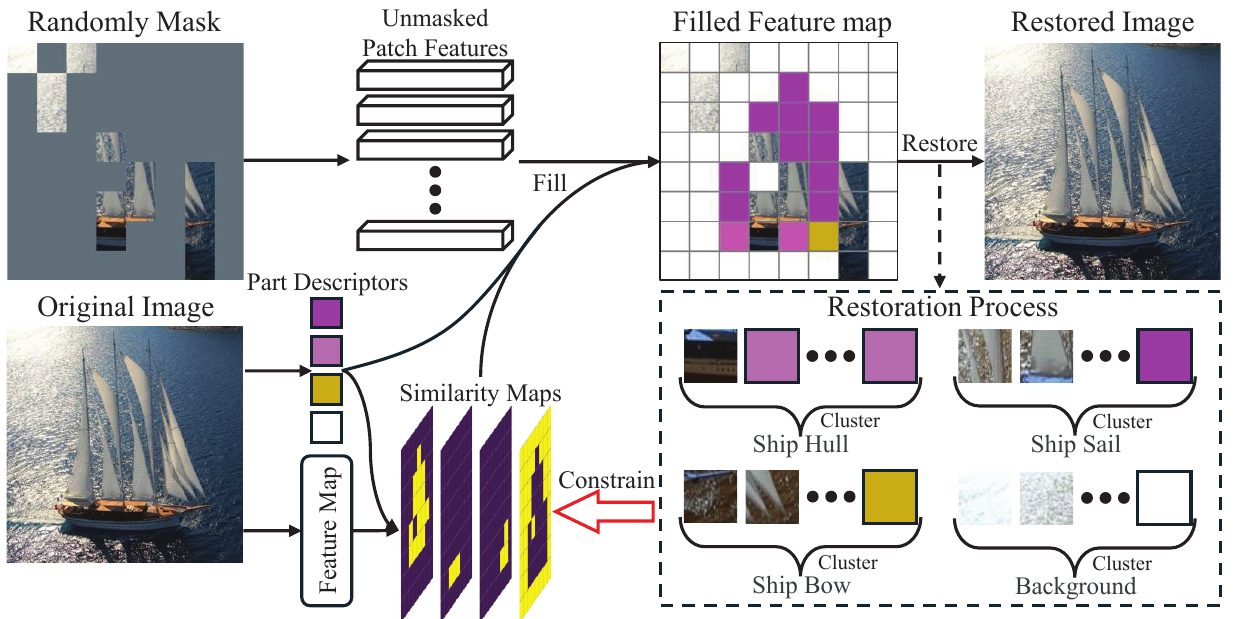}
	\caption{MPAE uses similarity maps between local image features and part descriptors to assign these descriptors to fill masked regions in a randomly masked version of the inputs. The restoration implicitly clusters the features of the restored patches that share similar appearance, further aligning low-level appearance in unmasked patches and high-level semantics of part descriptors. This drives similarity maps to closely match the actual part shapes.}
	\label{fig1}
\end{figure}

To eliminate the reliance on part-level labels, existing methods achieve part discovery mainly through two strategies: reconstruction \cite{ZhangLandmark, Scops, MotionCoPart, LiuDis, PartAssembly, GANSeg, Capsule, PartAttention} and clustering \cite{NIPSPart, DINO}. By directly reconstructing the input images using learned part descriptors, the descriptors are expected to align with the part shapes, allowing these part descriptors to further predict part masks. However, to discover meaningful parts, these descriptors are designed to learn high-level semantics rather than low-level appearance \cite{amir2021deep}, which prevents such reconstruction and alignment, especially in complex scenarios. Clustering-based methods use foreground masks to remove irrelevant regions and directly cluster foreground features, regardless of their actual appearance similarity. Some near-outlier features are identified as separate parts, leading to the discovery of unreasonable parts.

To address these limitations and extend unsupervised part discovery to more complex scenarios, we propose a novel paradigm, named Mask Part Autoencoder (MPAE). As shown in Fig. 1, instead of directly using part descriptors for image reconstruction, we first calculate the similarities between these descriptors and a feature map learned from the inpus. Based on the similarity maps, we assign these descriptors to fill the masked regions in a randomly masked version of inputs. By utilizing both filled part descriptors and unmasked patch features within the same part regions to generate image patches with similar appearances, these features are implicitly clustered together \cite{MAE_Matter}. The appearance features of unmasked patches (e.g., boundaries and textures) further align the high-level semantics of the part descriptors with the part shapes. Guided by part appearance, the similarity maps closely match the shapes of corresponding parts even in complex scenarios. Thus, they can be viewed as pixel-level masks of the discovered parts. 

Moreover, without using category labels, it is difficult for unsupervised methods to determine which object parts are present or absent in an image, because different categories consist of different types of parts. As a result, most existing methods \cite{Capsule, GANSeg, PartAssembly, LiuDis} are restricted to a single category and do not consider the absence of individual parts. However, it is still impossible for all parts to appear in every image, even within a single category, because objects can be occluded. Although \cite{PartAttention} achieves part discovery on a dataset with multiple categories, it still relies on category labels to determine which parts appear before part discovery. Therefore, we propose a looser constraints to enable MPAE to identify the presence of parts across various scenarios and categories in an unsupervised manner. Additionally, a more effective constraint is proposed to ensure that the MPAE discovers part-level masks, as the model can easily degrade to predicting only a single instance-level mask when part absence is allowed. This provides a foundation for tackling challenges posed by occlusion. As a result, these optimized constraints allow MPAE to leverage similar parts across different categories to achieve more robust part discovery without relying on any labels.

To evaluate the effectiveness of MPAE and our optimized constraints, extensive experiments are conducted on four widely used benchmarks, including two with multiple categories and two with a single category. The experimental results clearly demonstrate that MPAE robustly discovers meaningful object parts across diverse categories and achieves more competitive performance on all datasets than other state-of-the-art methods.

The main contributions of this paper can be summarized as followings:

\begin{itemize}
	\item We present a novel paradigm named Masked Part Autoencoder (MPAE), which utilizes low-level appearance features of the unmasked patches to align the part descriptors with actual part shapes. Consequently, MPAE discovers meaningful parts that closely match the object shapes.
	
	\item Several looser yet effective constraints are proposed to enable MPAE to identify the presence of parts across various scenarios and categories in an unsupervised manner. As a result, MPAE can leverage similar parts across various categories to achieve more robust part discovery.
	
	\item Extensive experiments are carried out on four widely-used benchmarks. The results show that our method can robustly discover reasonable object parts in complex scenarios and achieve highly competitive performance.
	
\end{itemize}

\label{sec:intro}

\section{Related works}

\subsection{Unsupervised / Self-supervised Learning}

Many works in natural language processing (NLP)~\cite{BERT, GPT} show unsupervised / self-supervised learning learns more effective features for downstream tasks than supervised learning. Inspired by their success, recent works~\cite{MOCO, MOCOV3, DINO, BYOL, MAE, Ziegler_self, DINOV2, Simaese_MAE, PARTICLE, DINO_Reg} also adopt unsupervised~/~self-supervised learning to improve the effectiveness of visual features. In addition to performance improvements, some new properties also emerge in Vision Transformer (ViT) \cite{VIT} with the use of unsupervised / self-supervised learning. Caron et al.~\cite{DINO} found using momentum encoder \cite{EMA} and multi-crop augmentation \cite{Multicrop} drives the ViT features from the same category to have high similarity. Based on this properties, Wang et al.~\cite{NormCut} and Wang et al.~\cite{CutLearn, VideoCut} achieve foreground segmentation, instance segmentation, and video instance segmentation, respectively, in an unsupervised manner by using normalized cut \cite{NCUT} to directly group self-similar regions. Although learned ViT features within the same part exhibit higher similarity~\cite{Ziegler_self, PARTICLE}, discovering object parts from images is still much more challenging than discovering instances. The main reason is that features from different parts also have very high similarity, resulting in less consistent semantics and boundaries in the discovered parts. Unfortunately, there are relatively few works that focus on using unsupervised / self-supervised features to discover meaningful and consistent object parts.

\subsection{Part Discovery}

Part discovery aims to identify meaningful object parts with consistent semantics and predict their pixel-level masks. Existing works mainly achieve part discovery through weakly-supervised or unsupervised learning. Weakly-supervised part discovery utilizes instance-level labels, such as category~\cite{HuangAttention, PDiscoNet, PDiscoFormer} and identity~\cite{Part_ReID, PLOT}, to discover relevant parts with pixel-level masks during training. Then, part-level features are extracted using these masks and further boost the performance of downstream tasks. Nevertheless, without aligning their the semantics from part embeddings with actual object appearances through reconstruction, their part masks do not follow the actual object boundaries well, making them insufficiently fine-grained.

Unsupervised part discovery is more challenging than weakly-supervised part discovery because no labels are provided during training. Early works discover meaningful object parts with rough part masks by factorizing deep features \cite{DEF} or spatial embeddings \cite{ULD}. Other methods \cite{amir2021deep, NIPSPart} directly cluster the foreground features across all images to discover meaningful parts. Nevertheless, the semantics and boundaries of the predicted part masks are not sufficiently stable due to appearance biases in each image. Most state-of-the-art methods \cite{ZhangLandmark, Scops, MotionCoPart, LiuDis, PartAssembly, Capsule, PartAttention} employ image reconstruction as a learning objective to align the part descriptors with the actual part shapes. Therefore, these part descriptors can further predict the part masks with consistent semantics. However, these descriptors are designed to capture high-level semantics rather than low-level appearance, which hinders reconstruction and alignment, ultimately degrading part discovery, especially in complex scenarios. Amir et al. \cite{amir2021deep} demonstrate that directly clustering self-supervised features in foreground regions can also discover meaningful parts. Nevertheless, they only consider feature similarity, disregarding actual appearance similarity. As a result, some near-outlier features are identified as independent parts, leading to unreasonably segmented parts. Therefore, robustly discovering meaningful parts in an unsupervised manner remains a significant challenge.

\label{sec:related works}
\section{Methodology}

\begin{figure*}[t!]
	\centering
	\includegraphics[width=0.92\linewidth]{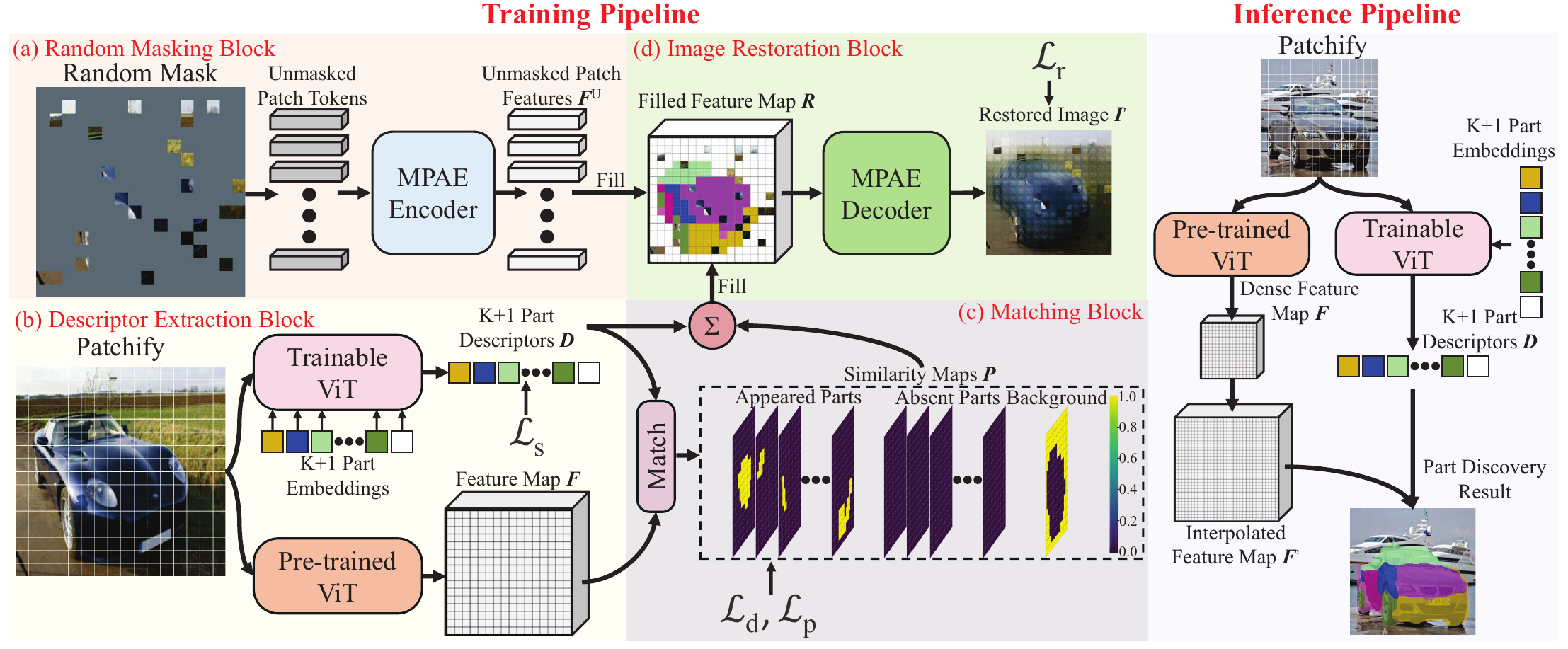}
	\caption{Training and inference pipelines of MPAE. \textbf{The training pipeline} consists of four blocks: \textbf{(a)} random masking block, \textbf{(b)} descriptor extraction block, \textbf{(c)} matching block, and \textbf{(d)} image restoration block. \textbf{(a)} randomly masks out a large proportion of patches and extracts patch features $\bm{F}^{\rm U}$ from the unmasked patches using MPAE encoder. The input image $\bm{I}$ is also fed into \textbf{(b)}, after being patchified, to extract part descriptors $\bm{D}$ and a feature map $\bm{F}$. By matching $\bm{D}$ with $\bm{F}$ using Hadamard product and a Softmax function, we generate a similarity map for each descriptor in \textbf{(c)}. Based on these similarity maps $\bm{P}$, we use $\bm{D}$ to fill the masked regionss and reconstruct a feature map $\bm{R}$. In \textbf{(d)}, $\bm{R}$ is further utilizes for image restoration, aligning the high-level semantics of the $\bm{D}$ with the corresponding part shapes. \textbf{The inference pipeline} directly employs $\bm{D}$ as mask prototypes to predict part-level masks from the Interpolated feature map $\bm{F^{\rm \prime}}$.}
	\label{fig2}
\end{figure*}

Unsupervised part discovery aims to learn a model that can parse all objects into a total of $K$ parts without any labels. Given an input image $\bm{I} \in \mathbb{R}^{H_{\rm I} \times W_{\rm I} \times 3}$, the learned model predicts pixel-level mask $\bm{M}_{\rm p} \in \{0, 1\}^{H_{\rm I} \times W_{\rm I} \times (K+1)}$ to assign each pixel into one of $K$ parts or background, where $(H_{\rm I}, W_{\rm I})$ is the size of the input image.

To achieve this goal, we present a novel paradigm, named \textbf{M}asked \textbf{P}art \textbf{A}uto\textbf{e}ncoder (MPAE), and further incorporate it with several optimized constraints. In what follows, we first detail the training and inference pipeline of MPAE, followed by an introduction to our optimized constraints and their corresponding functions.

\subsection{Masked Part Autoencoder}

\subsubsection{Training pipeline}

The overall training pipeline of MPAE is shown in Fig. 2. It consists of four blocks: (a) random masking block, (b) descriptor extraction block, (c) matching block, (d) image restoration block.

\textbf{Random masking block} first divides the input image $\bm{I}$ into patches of size $p$. The patchified image $\bm{I}^{\rm p}$ is denoted as $\bm{I}^{\rm p} \in \mathbb{R}^{\frac{H_{\rm I}}{p} \times \frac{W_{\rm I}}{p} \times 3p^2}$. A binary mask $\bm{M} = \{0, 1\}^{\frac{H_{\rm I}}{p} \times \frac{W_{\rm I}}{p}}$ is then generated to randomly mask out a specified proportion $r$ of the image patches. As MAE \cite{MAE}, these masked patches will not be used in the following training. Only visible patches are finally fed into the MPAE encoder to learn unmasked patch features $\bm{F}^{\rm U} \in \mathbb{R}^{\frac{H_{\rm I} \times W_{\rm I}}{p\times p}(1-r) \times C}$, where $C$ is the dimension of the unmasked patch feature. 

\textbf{Descriptor extraction block} is designed to learn part-level descriptors from input images. Specifically, as \cite{PartAttention}, we replace the class embedding of a classical ViT with $K + 1$ embeddings to extract $K + 1$ descriptors. By incorporating MPAE with appropriate constraints, each embedding learns to specialize in one potential part, adaptively aggregating image features to produce the corresponding part descriptor. The $K + 1$ part descriptors $\bm{D} \in \mathbb{R}^{(K+1) \times C}$ include $K$ descriptors for foreground parts and 1 for the background, where the descriptor dimension $C$ is the same as that of $\bm{F}^{\rm U}$.

To determine how to assign the sparse part descriptors $\bm{D}$ to fill the masked regions, we employ a frozen pre-trained ViT, followed by a trainable $1 \times 1$ convolutional layer, to extract a dense feature map $\bm{F} \in \mathbb{R}^{H_{\rm F} \times W_{\rm F} \times C}$, where $(H_{\rm F}, W_{\rm F})$ is the size of the feature map. By matching $\bm{F}$ with $\bm{D}$, we can obtain a similarity map for each part (as detailed in the following paragraphs). Based on the similarity maps, we can fill the masked patches using $\bm{D}$. Moreover, it enables PartFormer to leverage knowledge from the ViT, pre-trained through unsupervised or self-supervised learning, to discover more reasonable object parts.

\textbf{Matching block} calculates similarity maps $\bm{P}$ between $\bm{D}$ and $\bm{F}$, which can be defined as: 
\begin{equation}
	P_{i, j, k} = \frac{\exp{(\bm{D}_{k} \cdot \bm{F}_{i, j})}}{\sum_{k^\prime=1}^{K+1}\exp{(\bm{D}_{k^\prime} \cdot \bm{F}_{i, j})}},
\end{equation}
where $\bm{F}_{i, j}$ indicates the pixel-level feature at $(i, j)$ of $\bm{F}$, and $\bm{D}_k$ is the part descriptor for the $k$-th part, with $i \in \{1, \cdots, W_{\rm F}\}$, $j \in \{1, \cdots, H_{\rm F}\}$, and $k \in \{1, \cdots, K+1\}$. $P_{i, j, k}$ represents the normalized similarity between the $\bm{D}_k$ and $\bm{F}_{i, j}$. Based on the similarity map, the part descriptors are assigned to fill the masked regions and reform a dense feature map $\bm{R}$, which can be formulated as:
\begin{equation}
	\bm{R}_{i, j}=\left\{
	\begin{aligned}
		&\bm{F}^{\rm U}_{i,j} &, & M_{i,j}=0,\\
		&\sum^{K+1}_{k=1}\left(P_{i, j, k}\bm{D}_{k}\right) &,& M_{i,j}=1,
	\end{aligned}
	\right.
\end{equation}
where $\bm{R}_{i,j}$ is the feature vector of the filled feature map $\bm{R}$ at position $(i, j)$. $M_{i,j}$ is the value of the binary mask $\bm{M}$, indicating that the position $(i, j)$ is masked or not in random masking block. 

\textbf{Image restoration block} feeds $\bm{R}$ into the MPAE decoder to restore the masked patches, which can be written as:
\begin{equation}
	\mathcal{L}_{\rm r} = \frac{1}{2} \Vert \bm{I} - \bm{I}^\prime \Vert_{1} + \frac{1}{2} \Vert \Phi(\bm{I}) - \Phi(\bm{I}^\prime) \Vert_{1},
\end{equation}
where $\bm{I}^\prime$ is the restored image and $\Phi$ is a frozen VGG-19~\cite{VGG}. Directly using the L1 loss constraint for image restoration, as in~\cite{MAE}, can lead to structure deviations between the raw and restored images \cite{MAE}, which in turn causes the features in different parts to be poorly clustered, regardless of their actual appearance similarity. Therefore, in addition to using the L1 loss, we employ VGG-19 to encourage greater structural similarity between the restored image and the input image as the perceptual loss~\cite{Perceptual}.

The functions of this image restoration are threefold: 1) It retains low-level appearance features in MPAE through the unmasked patch features $\bm{F}^{\rm U}$, thereby compensating for the lack of low-level appearance features in the part descriptors $\bm{D}$. 2) It implicitly clusters $\bm{F}^{\rm U}$ and $\bm{D}$ within the same part region together in latent space by utilizing them to produce the image patches with similar appearance. Guided by low-level appearance features from $\bm{F}^{\rm U}$, the high-level semantics from $\bm{D}$ are aligned more accurately with the actual part shapes compared to directly using $\bm{D}$ from image reconstruction. 3) This alignment further constrains the similarity maps $\bm{P}$, encouraging them to more closely match the actual shapes of their corresponding parts, even in highly complex scenarios. Consequently, $\bm{P}$ can serve as pixel-level masks for the discovered parts.

\subsection{Learning Constraints}
To ensure robust part discovery, we incorporate MPAE with three types of optimized constraints: the presence constraint $\mathcal{L}_{\rm p}$, the semantic consistency constraint $\mathcal{L}_{\rm s}$, and the distribution constraint $\mathcal{L}_{\rm d}$.

\textbf{Presence constraint} $\mathcal{L}_{\rm p}  = \mathcal{L}_{\rm f} + \mathcal{L}_{\rm b}$, where $\mathcal{L}_{\rm f}$ and $\mathcal{L}_{\rm b}$ represent the foreground and background presence losses respectively. $\mathcal{L}_{\rm p}$ is directly applied to $\bm{P}$ to ensure the presence of $K$ foreground parts across the entire data and the presence of background in every image. 

$\mathcal{L}_{\rm f}$ divides each mini-batch into $N_{\rm g}$ mini-groups with $G$ samples. Then, $\mathcal{L}_{\rm f}$ can be calculated as:
\begin{equation}
	\mathcal{L}_{\rm f} = \frac{1}{N_{\rm g}}\sum_{n_{\rm g}=1}^{N_{\rm g}}(2-\frac{1}{G}\sum_{g=1}^{G}\max_{i, j, k}\bm{P}_{i, j, k}^{g, n_{\rm g}} -\frac{1}{K}\sum_{k=1}^{K}\max_{i,j, g} \bm{P}_{i, j, k}^{g, n_{\rm g}}),
\end{equation}
where $\bm{P}_{i, j, k}^{g, n_{\rm g}}$ is the similarity $\bm{P}_{i, j, k}$ of $g$-th sample in $n_{\rm g}$-th mini-group. Instead of requiring all parts to appear in each image as previous unsupervised methods, $\mathcal{L}_{\rm f}$ only encourages them to appear at least once in each mini-group and each image to contain at least one part. This looser constraint provides the basis for MPAE to identify the presence of parts across various scenarios and categories. The influence of the mini-group will be discussed in the \textbf{Appendix}.

$\mathcal{L}_{\rm b}$ encourages the background to appear near the boundaries of each image. Inspired by \cite{PDiscoFormer}, we define $\mathcal{L}_{\rm b}$ as:
\begin{equation}
	\mathcal{L}_{\rm b} = -\frac{1}{N_{\rm g}G}\sum_{n_{\rm g}=1}^{N_{\rm g}}\sum_{g=1}^{G}\log(\max_{i, j}d_{i, j}\bm{P}_{i, j, K+1}^{g, n_{\rm g}}),
\end{equation}
where $d_{i, j}$ is the distance from position $(i, j)$ to the image center. $d_{i, j}$ can be calculated as:
\begin{equation}
	d_{i, j} = 2\left(\frac{i - 1}{W_{\rm F} -1} - \frac{1}{2}\right)^2 + 2\left(\frac{j - 1}{H_{\rm F} -1} - \frac{1}{2}\right)^2.
\end{equation}
Incorporated with $\mathcal{L}_{\rm f}$ and $\mathcal{L}_{\rm b}$, our model also learns to distinguish the $K$ foreground parts and the background region.

\textbf{Semantic constraint} $\mathcal{L}_{\rm s}$: because $\mathcal{L}_{\rm p}$ only requires each image to contain at least one part, the model tends to simply segment out the entire foreground object, thereby degrading into an instance-discovery model. By encouraging each part descriptor to align only with the local features that exhibit very high similarity, $\mathcal{L}_{\rm s}$ ensures that the model decomposes the target into multiple object parts with consistent semantics. $\mathcal{L}_{\rm s}$ can be formulated as:
\begin{equation}
	\mathcal{L}_{\rm s}\!=\! -\!\frac{1}{K}\!\sum_{k=1}^{K}\log\frac{M^{k}e^{s\left(\cos\left(\theta_{\left(k, k\right)}+m\right)\right)}}{\!M^{k}e^{s\left(\cos\left(\theta_{\left(k, k\right)}+m\right)\right)}\!+\!\!\!\!\!\!\sum\limits_{t=1, t\neq k}^{K}\!\!\!\!\!M^{t}e^{s\cos\theta_{\left(t, k\right)}}},
\end{equation}
where $M_{k} \in \{0, 1\}$ represents the presence of the $k$-th part. When $\sum_{i=1}^{W_{\rm F}}\sum_{j=1}^{H_{\rm F}}P_{i, j, k} <= 0.001$, $M_{k}=0$, else, $M_{k}=1$. $m$ and $s$ are two hyperparameters. $\theta_{(k, t)}$ is the angle between the $k$-the part descriptor $\bm{D}_{k}$ and the mean feature vector of $t$-th part region $\bar{\bm{F}}_{t}$ in latent space. $\bar{\bm{F}}_{t}$ and $\cos(\theta_{(k, t)})$ are defined as:
\begin{equation}
	\bar{\bm{F}}_t = \frac{\sum_{i=1}^{W_{\rm I}}\sum_{j=1}^{H_{\rm I}}P_{i, j, t}\bm{F}_{i, j}}{\sum_{i=1}^{W_{\rm I}}\sum_{j=1}^{H_{\rm I}}P_{i, j, t}},
\end{equation}
\begin{equation}
	\cos(\theta_{(t, k)}) = \frac{\bm{D}_{k} \cdot \bar{\bm{F}}_{t}}{\Vert \bm{D}_{k}\Vert \Vert \bar{\bm{F}}_{t}\Vert}.
\end{equation}
As \cite{ArcFace}, $m$ enables $\mathcal{L}_{s}$ to still have a certain gradient value for further maximizing $\cos(\theta_{(k,k)})$ when $\theta_{(k,k)}$ is very small, and $s$ controls the smoothness of $\mathcal{L}_{s}$. By minimizing $\theta_{(k,k)}$ and maximizing $\theta_{(t,k)}$, the discovered parts across different objects have more consistent semantics. Moreover, the use of $M^k$ enables $\mathcal{L}_{s}$ to only consider the semantic consistency of those appeared parts, providing the basis to constrain the semantic consistency across multiple categories.

\textbf{Distribution constraint} $\mathcal{L}_{\rm d} = \mathcal{L}_{\rm v} + \mathcal{L}_{\rm e}$, where $\mathcal{L}_{\rm v}$ and $\mathcal{L}_{\rm e}$ are total variation loss \cite{TotalVariation} and entropy loss respectively. $\mathcal{L}_{\rm d}$ encourages the predicted probability map $\bm{P}$ to better follow the shapes of object parts.

$\mathcal{L}_{\rm v}$ minimizes the spatial gradient of the predicted $\bm{P}$, which can be formulated as:
\begin{equation}
	\mathcal{L}_{\rm v} = \frac{1}{H_{\rm F}W_{\rm F}}\sum_{k=1}^{K+1}\sum_{i=1}^{W_{\rm F}}\sum_{j=1}^{H_{\rm F}} \lvert \nabla P_{i, j, k} \rvert ,
\end{equation}
where $\nabla P_{i, j, k}$ represents the spatial image gradient at position $(i, j)$ on the probability map for the $k$-th part. The function of $\mathcal{L}_{\rm v}$ is the same as that of the concentration loss \cite{Scops}, encouraging pixels with high confidence for the same part to form a connected region. However, the concentration loss penalizes pixels that are far from the part center much more heavily than those that are close. This imbalance further causes the discovered parts to take on an approximately square shape. $\mathcal{L}_{\rm v}$ successfully eliminates this imbalance by calculating the gradient of $\bm{P}$, enabling the discovered parts to more accurately follow the shapes of objects, even when they are very complex.

$\mathcal{L}_{\rm e}$ ensures that the distribution at any position on $\bm{P}$ has high similarity in only one part descriptor by penalizing its entropy:
\begin{equation}
	\mathcal{L}_{\rm e} = -\frac{1}{K+1}\sum_{k=1}^{K+1}\sum_{i=1}^{W_{\rm F}}\sum_{j=1}^{H_{\rm F}}P_{i, j, k}\log\left(P_{i, j, k}\right).
\end{equation}
This drives the similarity map $\bm{P}$ to exhibit clear and stable boundaries, resulting in more precise part discovery.

\textbf{Overall training target}: by minimizing $\mathcal{L}=\mathcal{L}_{\rm r} + \lambda_{\rm p}\mathcal{L}_{\rm p} + \lambda_{\rm s}\mathcal{L}_{\rm s} + \lambda_{\rm d}\mathcal{L}_{\rm d}$, MPAE learns to discover meaningful parts in complex scenarios, where $\lambda_{\rm p}$, $\lambda_{\rm s}$, and $\lambda_{\rm d}$ are the weights of $\mathcal{L}_{\rm p}$, $\mathcal{L}_{\rm s}$, $\mathcal{L}_{\rm d}$ respectively.

\subsection{Inference pipeline}

The overall inference pipeline is shown on the right in Fig. 2. In this phase, we retain only PartFormer and the pre-trained ViT. However, the low-resolution of $\bm{F}$ is insufficient to produce high-resolution part masks. Therefore, we resize $\bm{F}$ to match the size of the input images using bilinear interpolation. By accurately aligning with part shapes with the guide of low-level appearance features, $\bm{D}$ can still accurately respond to the corresponding part regions in the $\bm{F}^{\prime}$ with much higher resolution. Thus, the similarity maps calculated from $\bm{D}$ and $\bm{F}^{\prime}$ can serve as high-resolution part masks for the parts discovered in an unsupervised manner.

\section{Experimental Setup}

\subsection{Datasets}

To comprehensively validate the effectiveness of MPAE, we evaluate it on four widely-used benchmarks: two with multiple categories (PartImageNet OOD dataset \cite{PartImageNet} and PartImageNet Segmentation dataset \cite{PartImageNet}) and two with a single category (CUB dataset~\cite{CUB} and CelebA dataset~\cite{CelebA}). More details about these four datasests and the implementation of our method are provided in \textbf{Appendix A.1} and \textbf{Appendix A.2} respectively.

\begin{table}[t!]
	\centering
	{\resizebox{\linewidth}{!}{
			\begin{tabular}{m{2.2cm}<{\centering}|m{1.2cm}<{\centering}|m{0.5cm}<{\centering}|m{1.1cm}<{\centering}m{1.1cm}<{\centering}|m{1.1cm}<{\centering}m{1.1cm}<{\centering}}
				\hline
				\multirow{2}{*}{Method} & 	\multirow{2}{*}{Label} & \multirow{2}{*}{K} & \multicolumn{2}{c|}{ PartImage\_O (\%)} & \multicolumn{2}{c}{PartImage\_S (\%)} \\ \cline{4-7}
				
				& & & NMI $\uparrow$ & ARI $\uparrow$ & NMI $\uparrow$ &  ARI $\uparrow$ \\ \hline
				\multicolumn{7}{c}{Weakly-supervised part discovery} \\ \hline
				\multirow{3}{*}{\shortstack{Huang et\\ al.~\cite{HuangAttention}}} & \multirow{3}{*}{\shortstack{Category\\ label}} & 8 & 5.88 & 1.53 & - & - \\
				& & 25 & 7.56 & 1.25 & - & - \\
				& & 50 & 10.19 & 1.05 & - & - \\ \hline
				\multirow{3}{*}{\shortstack{PDisco-\\Net~\cite{PDiscoNet}}} & \multirow{3}{*}{\shortstack{Category\\ label}} & 8 & 27.13 & 8.76 & 10.68 & 32.43 \\
				& & 25 & 32.41 & 10.69 & 18.44 & 40.70 \\
				& & 50 & 41.49 & 14.17 & 21.98 & 39.40 \\ \hline
				\multirow{3}{*}{\shortstack{PDiscoNet\\ + ViT-B/14}} & \multirow{3}{*}{\shortstack{Category\\ label}} & 8 & 19.28 & 34.72 & 16.68 & 32.01 \\
				& & 25 & 28.23 & 25.93 & \textbf{46.81} & 40.70 \\
				& & 50 & 29.48 & 28.68 & 34.13 & 39.40 \\ \hline
				\multirow{3}{*}{\shortstack{PDiscoFormer\\~\cite{PDiscoFormer}+ frozen\\ ViT-B/14$^\star$}} & \multirow{3}{*}{\shortstack{Category\\ label}} & 8 & 28.84 & 55.66 & - & - \\
				& & 25 & 43.36 & \textbf{62.82} & - & - \\
				& & 50 & 44.48 & 57.91 & - & - \\ \hline
				\multirow{3}{*}{\shortstack{PDiscoFormer\\+ ViT-B/14$^\dag$}} & \multirow{3}{*}{\shortstack{Category\\ label}} & 8 & 29.00 & 52.40 & 20.29 & 38.90 \\
				& & 25 & 44.71 & 59.27 & 43.18 & \textbf{64.61} \\
				& & 50 & \textbf{46.29} & 62.21 & 44.06 & 60.10 \\ \hline
				\multicolumn{7}{c}{Unsupervised part discovery} \\ \hline
				\multirow{3}{*}{DINO$^\star$ \cite{amir2021deep}} & \multirow{3}{*}{N/A} & 8 & 19.17 & 7.59 & - & - \\
				& & 25 & 31.46 & 14.16 & - & - \\
				& & 50 & 37.81 & 16.50 & - & - \\ \hline
				Xia et al.$^\star$~\cite{PartAttention} & N/A &$4N_{\rm c}$ & 27.85 & 40.77 & 25.17 & 37.50 \\ \hline
				\multirow{3}{*}{\shortstack{MPAE$^\star$ (ours)}} & \multirow{3}{*}{N/A} & 8 & 32.90 & 66.17 & 33.51 & 65.05 \\
				& &  25 & 39.28 & 68.87 & 37.22 & 68.04 \\
				& & 50 &  \textbf{53.65} & \textbf{74.22} & \textbf{55.10} & \textbf{73.52} \\ \hline
			\end{tabular}
	}}
	\caption{Quantitative comparisons with state-of-the-art methods on PartImageNet OOD and Segmentation datasets. $N_{\rm_c}$ represents the number of categories, which is 109 and 158 for the OOD and Segmentation datasets respectively. $\dag$= using a partially fine-tuned pretrained backbone, $^\star$=using a frozen pretrained backbone.}
	\label{Table1}
\end{table}

\subsection{Evaluation Metrics}
\textbf{Normalized Mutual information (NMI) and Adjusted Rand Index (ARI)} are primarily used to measure the similarity between two clusters. In this task, the NMI and ARI scores between the annotated and discovered parts directly reflect how closely the semantics of the discovered parts align with human conception.

\noindent\textbf{Normalized Mean Error (NME)}: NME is calculated using the centroids of the predicted parts and the annotated keypoints, representing the semantic consistency of the discovered parts across the entire test set.

\subsection{Comparisons with State-of-the-art Methods}
\textbf{PartImageNet OOD}: the quantitative results of MPAE and other state-of-the-art methods on PartImageNet OOD dataset are reported in Table 1, and some randomly chosen qualitative results are shown in Fig. 3. MPAE achieves highly competitive performance among both unsupervised and weakly-supervised methods. In the second row of Fig. 3, we observe that some small, round areas are identified as separate parts by DINO. These unreasonable parts result from directly clustering foreground features, which identifies certain near-outlier features as separate parts. By considering part appearance into clustering through our paradigm, we observe significant improvements in NMI of 71.62\%, 24.86\% and 41.89\% compared to DINO when $K=8, 25, 50$ respectively. Due to the complex scenarios, it is difficult to align high-level semantics from part descriptors with object shapes through reconstruction. As a result, the parts discovered by Xia et al. are often incomplete. Although PDiscoFormer and PDiscoNet employ classification loss for part discovery in a weakly supervised manner, they do not further align high-level semantics with object shapes. Consequently, their predicted masks do not match objects as closely as those predicted by MPAE. With the same backbone (frozen ViT-B/14), MPAE still outperforms them.

\begin{figure}[t!]
	\centering
	\includegraphics[width=\linewidth]{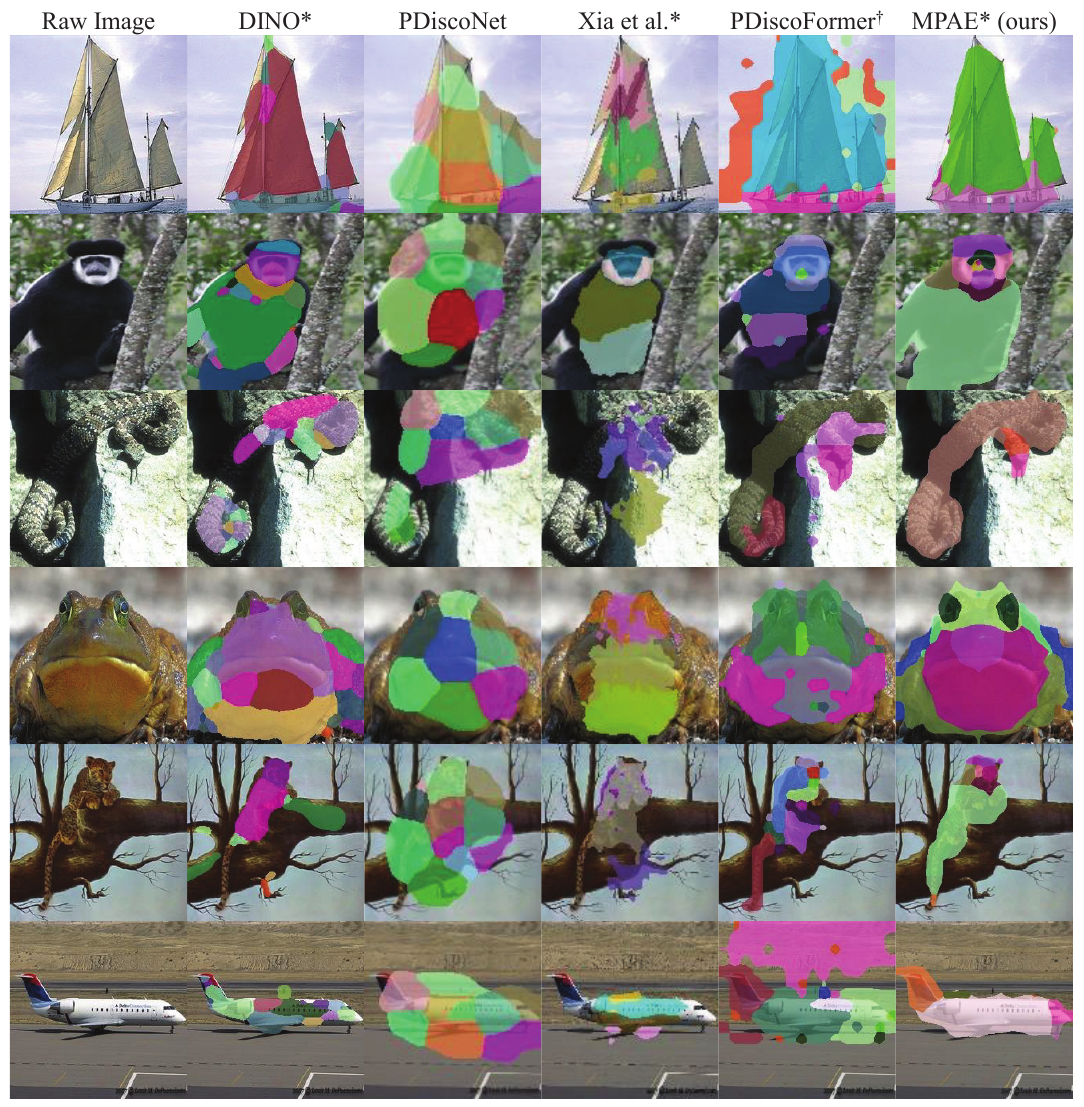}
	\caption{Some qualitative results of MPAE (ours) and other state-of-the-art unsupervised and weakly-supervised methods on PartImageNet OOD dataset ($K=50$).}
	\label{fig3}
\end{figure}

\begin{figure*}[t!]
	\centering
	\includegraphics[width=0.94\linewidth]{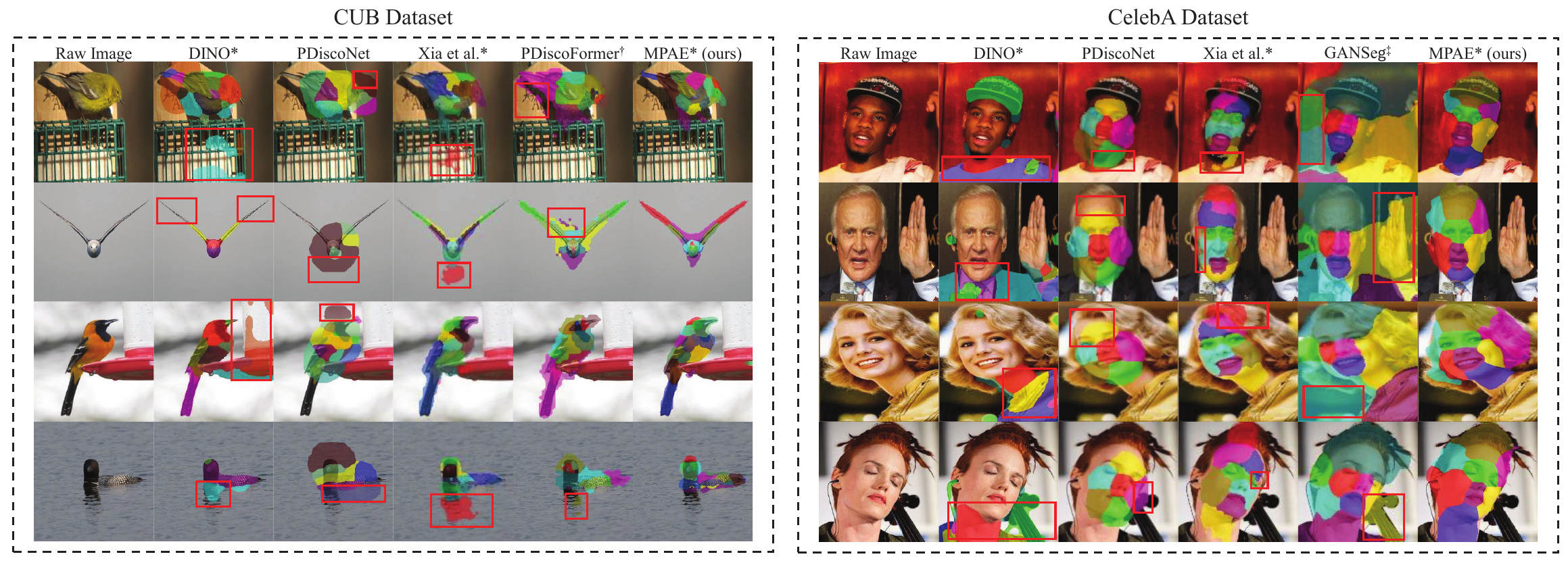}
	\caption{Some visualized part discovery results of MPAE (ours) and state-of-the-art methods on CUB dataset ($K=16$) and CelebA dataset ($K=8$). We use \textcolor[rgb]{1, 0, 0}{\textbf{red}} boxes to highlight the regions with poor part discovery results.}
	\label{fig4}
\end{figure*}

\begin{table*}[t!]
	\centering
	{\resizebox{0.88\linewidth}{!}{
			\begin{tabular}{m{4.4cm}<{\centering}|m{2.6cm}<{\centering}|m{0.8cm}<{\centering}m{0.8cm}<{\centering}m{0.8cm}<{\centering}|m{0.8cm}<{\centering}m{0.8cm}<{\centering}m{0.8cm}<{\centering}|m{0.8cm}<{\centering}m{0.8cm}<{\centering}m{0.8cm}<{\centering}}
				\hline
				\multirow{2}{*}{Method} & 	\multirow{2}{*}{Label} & \multicolumn{3}{c|}{$K=4$ (\%)} & \multicolumn{3}{c|}{$K=8$ (\%)} & \multicolumn{3}{c}{$K=16$ (\%)} \\ \cline{3-11}
				
				& & NME$\downarrow$ & NMI$\uparrow$ & ARI$\uparrow$ & NME$\downarrow$ & NMI$\uparrow$ &  ARI$\uparrow$  & NME$\downarrow$ & NMI$\uparrow$ &  ARI$\uparrow$\\ \hline
				\multicolumn{11}{c}{Weakly-supervised part discovery} \\ \hline
				Huang et al.~\cite{HuangAttention} & Species label& 11.51 & 29.74 & 14.04 & 11.60 & 35.72 & 15.90 & 12.60 & 43.92 & 21.01 \\
				PDiscoNet~\cite{PDiscoNet} & Species label & 9.12 & 37.82 & 15.26 & 8.52 & 50.08 & 26.96 & 7.64 & 56.87 & 38.05 \\
				Choudhury et al.~\cite{NIPSPart} & Foreground mask & 9.20 & 43.50 & 19.60 & - &  51.50 & 28.30 & - &- & - \\
				PDiscoNet + ViT-B/14 & Species label & 7.70 & 52.59 & \textbf{26.66} & 6.34 &  65.01 & 37.90 & 5.95 & 68.63 & 43.41 \\
				PDiscoFormer + ViT-B/14$^{\star}$~\cite{PDiscoFormer} & Species label &8.19 &  52.88 & 23.22 & \textbf{6.23} & 67.59 & 41.35 & 6.44 & 69.54 & 49.99 \\
				PDiscoFormer + ViT-B/14$^{\dag}$ & Species label & \textbf{7.41} & \textbf{58.13} & 25.11 & 6.54 & \textbf{69.87} & \textbf{42.76} & \textbf{5.74} & \textbf{73.38} & \textbf{55.83} \\ \hline
				\multicolumn{11}{c}{Unsupervised part discovery} \\ \hline
				DINO$^\star$~\cite{amir2021deep} & N/A & - & 31.18 & 11.21 & - & 47.21 & 19.76 & - & 50.57 & 26.14 \\
				Xia et al.$^\star$~\cite{PartAttention} & N/A & 9.92 & 45.22 & 19.40 & 8.58 & 52.42 & 26.42 &8.27& 53.52 & 33.95 \\
				MPAE$^\star$ (ours) & N/A & \textbf{7.75} & \textbf{58.93} & \textbf{33.73} & \textbf{5.53} & \textbf{68.24} & \textbf{45.85} & \textbf{5.14} & \textbf{79.00} & \textbf{66.56} \\ \hline
			\end{tabular}
	}}
	\caption{Quantitative comparisons with state-of-the-art methods on CUB. $\dag$= using partially fine-tuned backbone, $^\star$=using a frozen pretrained backbone.}
	\label{Table2}
\end{table*}

\begin{table*}[t!]
	\centering
	{\resizebox{0.88\linewidth}{!}{
			\begin{tabular}{m{4.4cm}<{\centering}|m{2.6cm}<{\centering}|m{0.8cm}<{\centering}m{0.8cm}<{\centering}m{0.8cm}<{\centering}|m{0.8cm}<{\centering}m{0.8cm}<{\centering}m{0.8cm}<{\centering}|m{0.8cm}<{\centering}m{0.8cm}<{\centering}m{0.8cm}<{\centering}}
				\hline
				\multirow{2}{*}{Method} & 	\multirow{2}{*}{Label} & \multicolumn{3}{c|}{$K=4$ (\%)} & \multicolumn{3}{c|}{$K=8$ (\%)} & \multicolumn{3}{c}{$K=16$ (\%)} \\ \cline{3-11}
				
				& & NME$\downarrow$ & NMI$\uparrow$ & ARI$\uparrow$ & NME$\downarrow$ & NMI$\uparrow$ &  ARI$\uparrow$  & NME$\downarrow$ & NMI$\uparrow$ &  ARI$\uparrow$\\ \hline
				\multicolumn{11}{c}{Weakly-supervised part discovery} \\ \hline
				Huang et al. \cite{HuangAttention} & Attribute label & \textbf{8.75} & 56.69 & 34.74 & \textbf{7.96} & 54.80 & 34.74 & \textbf{7.62} & 62.22 & 41.01 \\
				PDiscoNet~\cite{PDiscoNet} & Identity label & 11.11 & \textbf{75.97} & \textbf{69.53} & 9.82 & \textbf{62.61} & \textbf{51.89} & 9.46 & \textbf{77.43} & \textbf{70.58} \\ \hline
				\multicolumn{11}{c}{Unsupervised part discovery} \\ \hline
				DINO$^\star$~\cite{amir2021deep} & N/A & \textbf{11.36} & 1.38 & 0.01 & 10.74 & 1.12 & 0.01 & - & 3.29 & 0.06 \\
				GANSeg$^\ddag$~\cite{GANSeg} & N/A & 12.26 & 41.71 & 28.06  & \textbf{6.18} & \textbf{67.28} & \textbf{56.23} & - & - & - \\
				Xia et al.$^\star$~\cite{PartAttention} & N/A & 12.28 & 50.69 & 34.57 & 10.24 & 58.35 & 46.78 & - & - & - \\
				MPAE$^\star$ (ours) & N/A & 11.44 & \textbf{56.13} & \textbf{37.20} & 7.55 & 59.64 & 41.72 & \textbf{7.10} & \textbf{70.21} & \textbf{60.33} \\ \hline
				
			\end{tabular}
	}}
	\caption{Quantitative comparisons with state-of-the-art methods on CelebA. $^\star$=using a frozen pretrained backbone, $\ddag$=using a multi-stage self-training strategy.}
	\label{Table3}
\end{table*}

\noindent\textbf{PartImageNet Segmentation} increases category number $N_{\rm C}$ from 109 to 158, which means larger variance in object appearance and shape. Therefore, as shown in Table 1, the metrics in both NMI and ARI of PDiscoFormer and Xia et al. degrade slightly. However, by using image restoration to constrain the learning and by exploring part similarities across different categories, MPAE successfully maintains strong robustness on this more challenging dataset. Compared to PDiscoFormer, MPAE still achieves an impressive improvement of 65.16\% and 25.06\% respectively in NMI when $K$=8 and $50$, even though PDiscoFormers employ both partially fine-tuned backbone and category labels.

\noindent\textbf{CUB}: MPAE also achieves highly competitive performance on CUB (Table 2 and Fig. 4), demonstrating its effectiveness in part discovery within a single category. By implicitly clustering patch features with similar appearances and allowing for the absence of specific parts, MPAE can robustly discover meaningful bird parts in cases involving shadows, self-occlusion, and reflections (see first, second and fourth columns in Fig. 4). With the \textit{same frozen} pretrained ViT-B/14, MPAE outperforms PDiscoFormer in all metrics, despite not using any labels for training. Compared to PDiscoFormer with a partially fine-tuned ViT-B/14, MPAE still outperforms it by 10.45\%, 7.66\% and 19.22\% in NME, NMI and ARI respectively when $K=16$.

\noindent\textbf{CelebA}: as shown in Fig. 4, the masks predicted by MPAE accurately follow the entire facial region, resulting in significantly better qualitative results. Despite less challenging scenarios, the masks predicted by Xia et al. remain incomplete and fail to cover the entire face region. Other methods also misidentify some background regions as foreground parts. However, CelebA provides only five keypoints near the eyes, nose, and mouth for metric calculation, which is insufficient to reflect the improvements achieved by MPAE, as presented in Table 3.

\begin{figure}[t!]
	\centering
	\includegraphics[width=\linewidth]{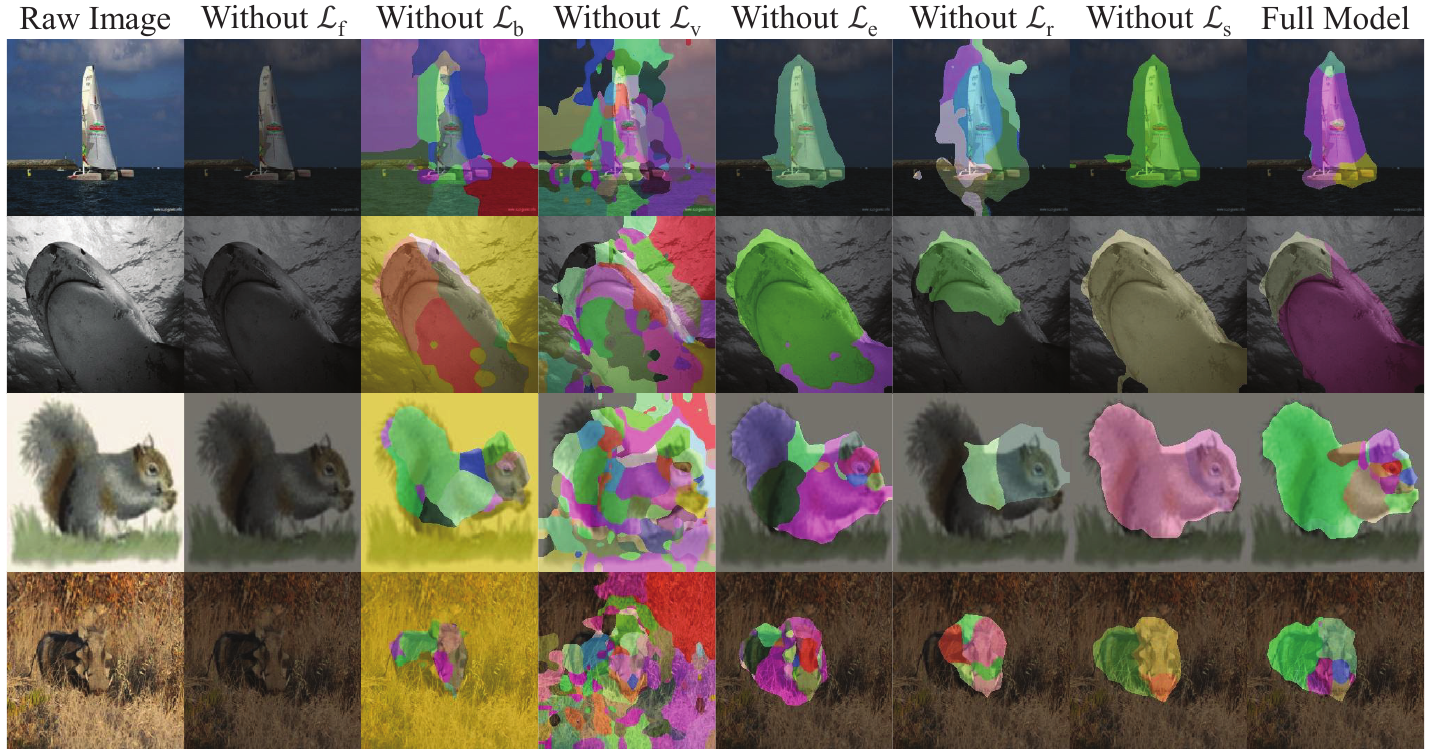}
	\caption{Qualitative results from the ablation studies of the loss functions on PartImageNet Segmentation dataset ($K=50$). From left to right: (1) raw images, models trained without (2) $\mathcal{L}_{\rm f}$, (3) $\mathcal{L}_{\rm b}$, (4) $\mathcal{L}_{\rm v}$, (5) $\mathcal{L}_{\rm e}$, (6) $\mathcal{L}_{\rm r}$, (7)$ \mathcal{L}_{\rm s}$ and (8) nothing (full model).}
	\label{fig5}
\end{figure}

\begin{table}[t!]
	\centering
	{\resizebox{\linewidth}{!}{
			\begin{tabular}{m{2.2cm}<{\centering}|m{1.0cm}<{\centering}m{1.0cm}<{\centering}|m{1.0cm}<{\centering}m{1.0cm}<{\centering}|m{1.0cm}<{\centering}m{1.0cm}<{\centering}}
				\hline
				\multirow{2}{*}{Method} & \multicolumn{2}{c|}{$K=8$ (\%)} & \multicolumn{2}{c|}{$K=25$ (\%)} & \multicolumn{2}{c}{$K=50$ (\%)} \\ \cline{2-7}
				& NMI$\uparrow$ & ARI$\uparrow$ & NMI$\uparrow$ & ARI$\uparrow$ & NMI$\uparrow$ & ARI$\uparrow$ \\ \hline
				Without $\mathcal{L}_{\rm f}$ & 0.00 & 0.00 & 0.00 & 0.00 & 0.00 & 0.00 \\
				Without $\mathcal{L}_{\rm b}$ & 30.91 & 73.20 & 33.67 & 67.85 & 39.36 & 55.24 \\
				Without $\mathcal{L}_{\rm v}$ & 18.30 & 20.62 & 13.52 & 4.54 & 14.37 & 4.08 \\
				Without $\mathcal{L}_{\rm e}$ & 31.21 & 63.27 & 27.48 & 54.93 & 52.51 & 73.08 \\
				Without $\mathcal{L}_{\rm r}$ & 25.60 & 57.39 & 31.53 & 63.39 & 22.32 & 49.73 \\
				Without $\mathcal{L}_{\rm s}$ & 55.69 & 76.54 & 57.70 & 76.97 & 56.42 & 74.03 \\ \hline
				Full model & 33.51 & 65.05 & 37.22 & 68.04 & 55.10 & 73.52 \\ \hline
			\end{tabular}
}}
\caption{Performance comparisons of MPAE without different loss functions on the PartImageNet Segmentation dataset in the setting of $K=8, 25, 50$.}
\label{Table4}
\end{table}

\subsection{Ablation Studies}
\noindent\textbf{Influence of loss functions}: to study the effectiveness of the loss functions used in MPAE, we conduct ablation studies on PartImageNet Segmentation dataset. The quantitative and qualitative results are shown in Table 4 and Fig. 5 respectively. \textbf{Foreground loss} $\mathcal{L}_{\rm f}$ and \textbf{background loss} $\mathcal{L}_{\rm b}$ are crucial for preventing the model from identifying all pixels as background or foreground. Otherwise, the learned model converges to a trivial solution. Without the prior knowledge provided by \textbf{Total variation loss} $\mathcal{L}_{\rm v}$, the pixels identified as foreground regions cannot form connected and concentrated parts. \textbf{Entropy loss} $\mathcal{L}_{\rm e}$ leads to more stable and smooth part boundaries by enforcing that each pixel highly responds to a unique part descriptor. The \textbf{absence of reconstruction loss} $\mathcal{L}_{\rm r}$ results in a failure to align part descriptors with actual part shapes. Consequently, the predicted masks contain only rough regions without clear part boundaries. $\textbf{Without semantic constraint}$ $\mathcal{L}_{\rm s}$, the model identifies the regions with less semantic similarity as the same part. In some cases, the learned model even degrades into a foreground segmentation model. Although this degradation may improve the numerical metrics in NMI and ARI, the learned model cannot achieve the goal of part discovery. Therefore, all constraints used in MPAE are essential for learning a robust model for unsupervised part discovery.

\begin{table}[t!]
	\centering
	{\resizebox{\linewidth}{!}{
			\begin{tabular}{m{2.4cm}<{\centering}|m{0.6cm}<{\centering}m{0.6cm}<{\centering}m{0.6cm}<{\centering}m{0.6cm}<{\centering}m{0.6cm}<{\centering}m{0.6cm}<{\centering}m{0.6cm}<{\centering}}
				\hline
				Masking Ratio $r$ & 50\% & 75\% & 80\% & 85\% & 90\% & 95\% & 100\% \\ \hline
				NMI (\%) $\uparrow$ & 45.50 & 50.11 & 52.20 & 52.70 & \textbf{55.10} & 54.63 & 43.39 \\
				ARI (\%) $\uparrow$ & 70.25 & 70.89 & 70.90 & 71.85 & \textbf{73.52} & 73.23 & 41.29 \\ \hline
			\end{tabular}
	}}
	\caption{Performance comparisons of MPAE with different masking ratio $r$ on the PartImageNet Segmentation dataset in the setting of $K=50$.}
	\label{Table5}
\end{table}

\textbf{Influence of the masking ratio $r$}: the quantitative results of MPAE with different masking ratios $r$ are reported in Table 5. When $r<90\%$, a higher $r$ results in more features in the filled feature map $\bm{R}$ comes from part descriptors $\bm{D}$. Therefore, by aligning the part descriptors $\bm{D}$ with object part shapes, a stronger constraint can be applied to the similarity maps $\bm{P}$, requiring them to better match the actual part shapes. As a results, we observe an improvement in performance. When $r>90\%$, a higher $r$ results in insufficient unmasked patches containing low-level appearance information to guide the alignment between descriptors and part shapes. Therefore, the optimal masking ratio $r$ is 90\%.

\begin{table}[t!]
	\centering
	{\resizebox{\linewidth}{!}{
	\begin{tabular}{m{4.6cm}<{\centering}|m{1.6cm}<{\centering}m{1.6cm}<{\centering}}
		\hline
		Change from MPAE model: &  NMI (\%) $\uparrow$ & ARI (\%)$\uparrow$ \\ \hline
		$\mathcal{L}_{\rm f}$ $\rightarrow$ Area loss in \cite{PartAttention}  & 1.90 & -0.14  \\
		$\mathcal{L}_{\rm p}$ $\rightarrow$ Presence loss in \cite{PDiscoFormer} & 42.45 & 60.24 \\
		$\mathcal{L}_{\rm d}$ $\rightarrow$ Concentration loss in \cite{Scops} & 17.29 & 18.74 \\
		Part number $K$ $\rightarrow$ $4N_{\rm c}$ & 47.59 & 70.71 \\ \hline
		MPAE (our best model) & \textbf{55.10} & \textbf{73.52} \\ \hline
		
	\end{tabular}
	}}
	\caption{Comparisons of our optimized losses and designs with previous ones on the PartImageNet Segmentation dataset in the setting of $K=50$.}
	\label{Table6}
\end{table}

\textbf{Improvements of our optimized constraints and designs}: The \textbf{area loss in \cite{PartAttention}} requires all parts to appear in each image. But, it is impossible for every object to contain all 50 parts. Consequently, $\mathcal{L}_{\rm a}$ leads to a collapse during training. The image classification loss in \cite{PDiscoFormer} ensures that each image to have at least one discovered part, as the embedding for classification cannot be a zero vector. Without any labels, \textbf{presence loss in \cite{PDiscoFormer}} cannot guarantee that unsupervised methods discover at least one part in each image. Therefore, some images may not have any discovered parts, which further leads to a degradation in metrics. The \textbf{concentration loss in \cite{Scops}} penalizes pixels that are farther from the part center more heavily, which prevents the predicted parts from accurately following their actual shapes, whereas $\mathcal{L}_{\rm d}$ addresses this imbalance. As \cite{PartAttention}, we also implement a model that simply decomposes each category into four parts without exploring their semantic similarities ($\bm{K}$ $\bm{\rightarrow}$ $\bm{4N_{\rm c}}$). Consequently, the number of training samples for each part significantly decreases, leading to reduced robustness.

\textbf{More ablation studies} and \textbf{visualized results} are included in the \textbf{Appendix}.

\section{Conclusion}
In this paper, we present a novel paradigm (MPAE) with looser yet effective constraints for more robust unsupervised part discovery. By restoring masked image patches using the corresponding part descriptors, MPAE effectively aligns their high-level semantics with the part shapes under the guidance of low-level appearance features from unmasked patches. As a result, MPAE predicts masks that closely match the actual shapes of corresponding parts. The optimized constraints enable MPAE to identify the presence of parts across various scenarios and categories in an unsupervised manner, effectively addressing challenges posed by occlusion and diverse object categories.

{
    \small
    \bibliographystyle{ieeenat_fullname}
    \bibliography{main.bib}
}
\newpage

\onecolumn

\section*{Appendix A.1 Dataset Details}
\label{sec:Appendix.A.1}

\textbf{PartImageNet OOD dataset}. We use the OOD variant of PartImageNet \cite{PartImageNet} to validate MPAE, following the setting used in \cite{PDiscoFormer, PDiscoNet}. This variant comprises 110 classes distributed across 11 super-classes, including 14,865 samples for training and 1,658 samples for testing. Each sample is annotated with pixel-level part masks.

\noindent\textbf{PartImageNet Segmentation dataset}. This dataset is another variant of PartImageNet \cite{PartImageNet}. Compared to the OOD variant, it is more challenging because the number of categories increases from 109 to 158, comprising 20,457 images for training and 2,405 images for testing.

\noindent\textbf{CelebA dataset}. CelebA dataset \cite{CelebA} contains 200,000 unaligned face images with 5 labeled keypoints, representing the eye centers, the tip of the nose, and the corners of the mouth, for 10,000 different identities. Following the setting in \cite{Scops}, we retain the images where face covers more than 30\% of the area, resulting in 45,609 images for training, 5,397 images for validation and 283 images for testing. This ensures the face to be the salient object in each image for subsequent part discovery.

\noindent\textbf{CUB dataset}. CUB dataset \cite{CUB} consists of 200 different bird species with 5,994 images for training and 5,794 images for testing. Each image is annotated with 15 keypoints and their visibility, representing 15 different bird parts.

\section*{Appendix A.2 Implementation Details}
\label{sec:Appendix.A.2}
As in \cite{HuangAttention, PDiscoNet}, we set the input size to $448 \times 448$ for CUB dataset and $224 \times 224$ for other datasets to ensure fair comparisons. For the datasets with multiple categories (PartImageNet OOD and PartImageNet Segmentation), the mini-group size is set to 64. For the datasets with single category, the mini-group size is set to 8. All models employ a frozen ViT-B/14 pre-trained using DINO v2 \cite{DINOV2} and register tokens \cite{DINO_Reg} to extract dense feature maps. Other parts of MPAE are trainable and are optimized using Adam optimizer \cite{Adam}. The learning rate, batch size, and feature dimension $C$ are set to $5 \times 10^{-3}$, 64 and 256 respectively. The number of both MPAE encoder layers and decoder layers is set to 2. In all experiments, $\lambda_{\rm p}$ is set to 1.0, $\lambda_{\rm d}$ to 0.5, and $\lambda_{\rm s}$ to 0.25. Referring to \cite{PartAttention}, we set $s$ and $m$ to 20 and 0.5, respectively, to ensure that each part descriptor is well aligned with the pixel-level features that have very high similarity.

\section*{Appendix B.1 Influence of Structural Difference Penalty Component of $\mathcal{L}_{\rm r}$}

In addition to minimizing the differences between masked image patches and their corresponding restored patches, we employ a frozen pretrained VGG-19 \cite{VGG} to penalize the structural difference between the masked and restored images. The quantitative and qualitative impact of this structural difference penalty component is demonstrated in Table 7 and Fig. 6 respectively in Appendix B.1. This penalty plays an essential role in the training of MPAE, increasing the NMI and ARI metrics from 19.65 and 49.72 to 55.10 and 73.52 respectively. From Fig. 6, we observe that the part discovery results are highly consistent with the images restored using their learned part descriptors: restored patches with similar views tend to be identified as the same part. This further supports the conclusion that MPAE implicitly clusters the filled part descriptors and unmasked patch features within the same part regions by utilizing them to generate image patches with similar appearances. Consequently, the low-level appearance features of the unmasked patches further align the high-level semantics of the part descriptors with the corresponding part shapes. Without the structural penalty, significant structural deviations can be observed between the input images and the restored images. This further results in a misalignment between the part descriptors and the shapes of their corresponding parts, as well as similarity maps that do not closely follow the part boundaries. Consequently, the MPAE fails to discover meaningful parts with consistent semantics, resulting in performance degradation in all metrics.

\begin{table}[H]
	\centering
	\begin{tabular}{m{3.0cm}<{\centering}|m{1.0cm}<{\centering}m{1.0cm}<{\centering}}
		\hline
		Structural Difference Penalty Component & With & Without \\ \hline
		NMI (\%) $\uparrow$ & \textbf{55.10} & 19.65  \\
		ARI (\%) $\uparrow$ & \textbf{73.52} & 49.72 \\ \hline
	\end{tabular}
	\caption{Performance comparisons of MPAE with/without the structural difference penalty component of $\mathcal{L}_{\rm r}$ on the PartImageNet Segmentation dataset in the setting of $K=50$.}
	\label{Table7}
\end{table}

\begin{figure}[H]
	\centering
	\includegraphics[width=0.8\linewidth]{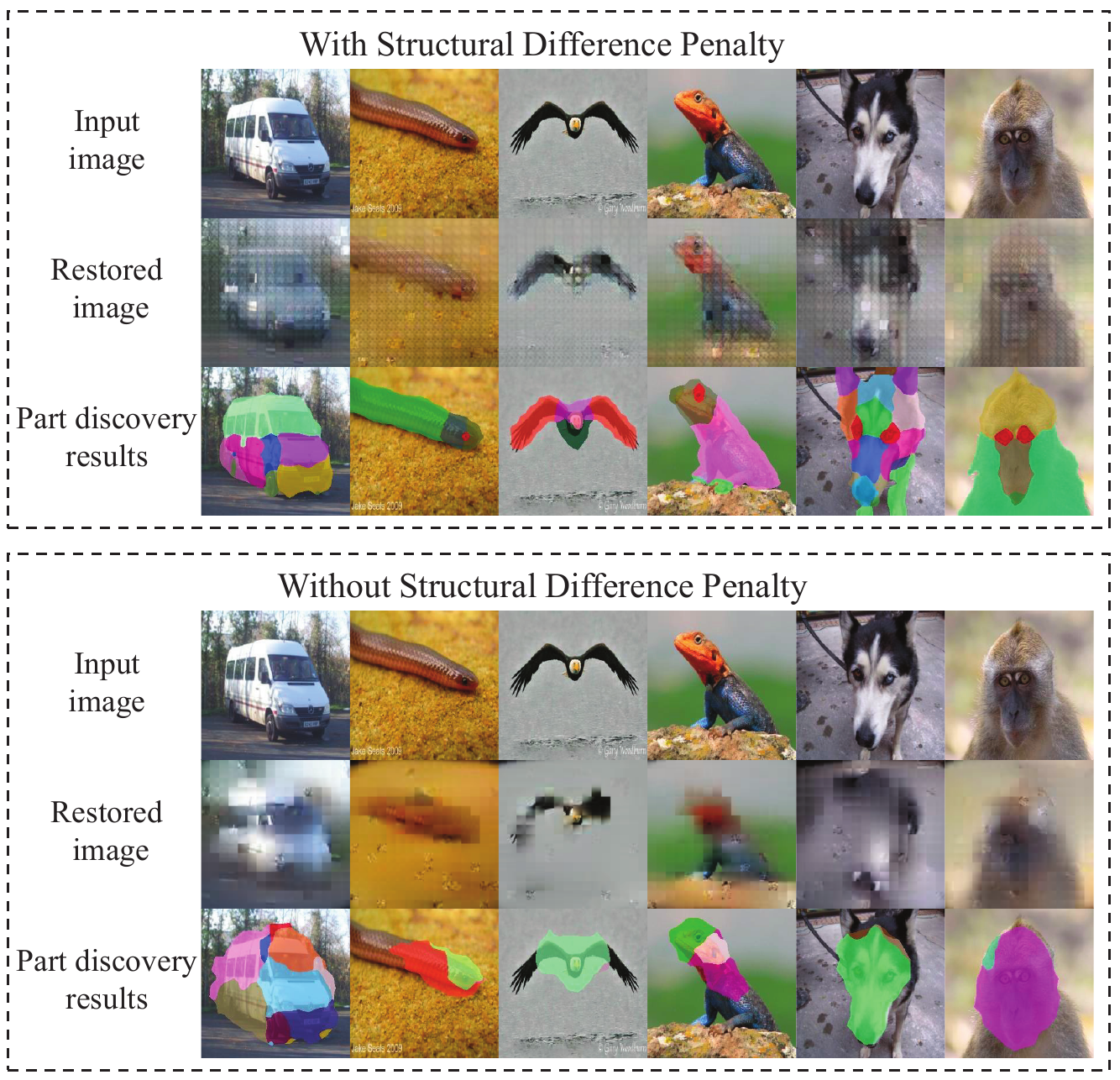}
	\caption{Some qualitative results of image restoration and part discovery predicted by the MPAE trained with/without structural difference penalty component in $\mathcal{L}_{\rm r}$.}
	\label{fig6}
\end{figure}

\section*{Appendix B.2 Influence of Encoder Layer Number}

\begin{table}[H]
	\centering
		\begin{tabular}{m{2.4cm}<{\centering}|m{0.6cm}<{\centering}m{0.6cm}<{\centering}m{0.6cm}<{\centering}m{0.6cm}<{\centering}}
			\hline
			MPAE encoder layer number & 1 & 2 & 4 & 6 \\ \hline
			NMI (\%) $\uparrow$ & 50.71 & 55.10 & 55.12 & \textbf{55.43} \\
			ARI (\%) $\uparrow$ & 70.86 & 73.52 & \textbf{73.74} & 73.59 \\ \hline
		\end{tabular}
	\caption{Performance comparison of MPAE with different number of encoder layers on the PartImageNet Segmentation dataset in the setting of $K=50$. The number of MPAE decoder layers is fixed at 2.}
	\label{Table8}
\end{table}

We conduct ablation studies on PartImageNet Segmentation dataset ($K=50$) to investigate the influence of the number of MPAE encoder layers. The results are reported in Table 8 in Appendix B.2. When the number of MPAE encoder layers increases from 1 to 2, the performance of MPAE improves from 50.71 to 55.10 in NMI. This is mainly because the expressive ability of the MPAE encoder with only a single layer is insufficient to effectively encode the appearance of unmasked patches in the latent space. Consequently, the high-level semantics from part descriptors and the low-level appearances are not well aligned, leading to performance degradation. Nevertheless, when the number of encoder layers exceeds two, the MPAE encoder can coherently encode the appearance of unmasked patches in the latent space. Therefore, the improvement gained from increasing the number of encoder layers is relatively slight.

\section*{Appendix B.3 Influence of Decoder Layer Number}

\begin{table}[H]
	\centering
	\begin{tabular}{m{2.4cm}<{\centering}|m{0.6cm}<{\centering}m{0.6cm}<{\centering}m{0.6cm}<{\centering}m{0.6cm}<{\centering}}
		\hline
		MPAE Decoder layer number & 1 & 2 & 4 & 6 \\ \hline
		NMI (\%) $\uparrow$ & 54.27 & \textbf{55.10} & 53.89 & 52.69 \\
		ARI (\%) $\uparrow$ & 73.32 & \textbf{73.52} & 72.43 & 71.81 \\ \hline
	\end{tabular}
	\caption{Performance comparison of MPAE with different number of decoder layers on the PartImageNet Segmentation dataset in the setting of $K=50$. The number of MPAE encoder layers is fixed at 2.}
	\label{Table9}
\end{table}

We also conduct ablation studies on PartImageNet Segmentation dataset ($K=50$) to investigate the influence of the number of MPAE decoder layers. The results are reported in Table 9 in Appendix B.3. Similarly, when the number of MPAE decoder layers increases from 1 to 2, we observe improvements in all metrics. The primary reason is that increasing the number of decoder layers from one to two improves image restoration results. Consequently, high-level semantics from part descriptors become better aligned with the shapes of their corresponding parts. However, when the number of decoder layers exceeds 2, we observe slight performance degradation. This is because more decoder layers encourages the model to rely more on the unmasked patch features for image restoration rather than part descriptors. As a result, the features within the same part region on the filled feature map $\bm{R}$ are not well aligned with the part shapes.

\section*{Appendix B.4 Influence of $\lambda_{\rm d}$, $\lambda_{\rm p}$ and $\lambda_{\rm s}$}

\begin{table}[H]
	{\resizebox{0.32\linewidth}{!}{
		\centering
		\begin{tabular}{m{1.2cm}<{\centering}|m{0.7cm}<{\centering}m{0.7cm}<{\centering}m{0.7cm}<{\centering}m{0.7cm}<{\centering}m{0.7cm}<{\centering}}
		\hline
		$\lambda_{\rm d}$ & 0.3 & 0.4 & 0.5 & 0.6 & 0.7 \\
		\hline
		NMI$\uparrow$(\%) & 35.88 & 53.21 & 55.10 & 55.45 & 55.36 \\
		ARI$\uparrow$(\%) & 31.39 & 72.89 & 73.52 & 74.85 & 73.93 \\
		\hline
		\end{tabular}
	}}
	\hfill
	{\resizebox{0.32\linewidth}{!}{
		\centering
		\begin{tabular}{m{1.2cm}<{\centering}|m{0.7cm}<{\centering}m{0.7cm}<{\centering}m{0.7cm}<{\centering}m{0.7cm}<{\centering}m{0.7cm}<{\centering}}
		\hline
		$\lambda_{\rm p}$ & 0.50 & 0.75 & 1.00 & 1.25 & 1.50 \\
		\hline
		NMI$\uparrow$(\%) & 53.50 & 55.55 & 55.10 & 53.46 & 52.68 \\
		ARI$\uparrow$(\%) & 74.62 & 73.61 & 73.52 & 73.08 & 72.45 \\
		\hline
		\end{tabular}
		}}
	\hfill
	{\resizebox{0.32\linewidth}{!}{
		\centering
		\begin{tabular}{m{1.2cm}<{\centering}|m{0.6cm}<{\centering}m{0.6cm}<{\centering}m{0.6cm}<{\centering}m{0.6cm}<{\centering}m{0.6cm}<{\centering}}
			\hline
			$\lambda_{\rm s}$ & 0.15 & 0.20 & 0.25 & 0.30 & 0.35 \\
			\hline
			NMI$\uparrow$(\%) & 55.45 & 55.23 & 55.10 & 53.00 & 51.28 \\
			ARI$\uparrow$(\%) & 74.59 & 75.48 & 73.52 & 69.34 & 68.75 \\
			\hline
			\end{tabular}
	}}
	\caption{Influence of $\lambda_{\rm d}$, $\lambda_{\rm p}$ and $\lambda_{\rm s}$ on PartImageNet Segmentation ($K=50$)}
	\label{Table10}
\end{table}

We carry out ablation studies on PartImageNet Segmentation dataset ($K=50$) to further investigate the influence of hyperparameters $\lambda_{\rm d}$, $\lambda_{\rm p}$ and $\lambda_{\rm s}$, as shown in Table 10 in Appendix B.4. Across a wide range of $\lambda_{\rm d}$, $\lambda_{\rm p}$ and $\lambda_{\rm s}$, MPAE consistently maintains comparable performance, indicating that it is not sensitive to the selection of hyperparameters. Moreover, we keep the hyperparameters fixed across all datasets in our paper, and MPAE still achieves competitive performance compared to other state-of-the-art methods, illustrating its robustness.

\section*{Appendix B.5 Influence of Different Self-supervised Pretrained Backbones}

\begin{table}[H]
	\centering
	\begin{tabular}{m{3.2cm}<{\centering}|m{1.0cm}<{\centering}m{1.0cm}<{\centering}|m{1.0cm}<{\centering}m{1.0cm}<{\centering}|m{1.0cm}<{\centering}m{1.0cm}<{\centering}}
	\hline
	\multirow{2}{*}{Backbone} & \multicolumn{2}{c|}{K=8 (\%)} & \multicolumn{2}{c|}{K=25 (\%)} & \multicolumn{2}{c}{K=50 (\%)} \\ \cline{2-7}
		& NMI & ARI & NMI & ARI & NMI & ARI \\ \hline
		DINO v1 \cite{DINO} & 26.86 & 63.70 & 32.89 & 69.75 & 39.12 & 70.08 \\
		DINO v2 \cite{DINOV2} with \cite{DINO_Reg} & \textbf{32.90} & \textbf{66.17} & \textbf{39.28} & \textbf{68.87} & \textbf{53.65} & \textbf{74.22} \\  \hline
	\end{tabular}
	\caption{Performance comparison of MPAE with different self-supervised pretrained backbones on the PartImageNet OOD dataset in the setting of $K=50$. The number of MPAE encoder and decoder layers is fixed at 2.}
	\label{Table11}
\end{table}

We implement MPAE with DINO v1 (ViT-S/16) on PartImage OOD, and the results are reported in Table 11 in Appendix B.5. Compared to direct clustering \cite{amir2021deep} and Xia et al. \cite{PartAttention}, which use similar backbone (DINO v1, ViT-S/8), our MPAE with DINO v1 still outperforms them by a significant margin, demonstrating the effectiveness of our method. However, the features produced by DINO V1 are not as fine-grained as those produced by DINO v2. Consequently, MPAE with DINO V1 fails to outperform MPAE with DINO v2.

\section*{Appendix B.6 Comparison with Supervised Pretrained Backbones}

\begin{table}[H]
	\centering
	\begin{tabular}{m{2.4cm}<{\centering}|m{1.8cm}<{\centering}m{1.8cm}<{\centering}m{1.8cm}<{\centering}}
		\hline
		Backbone & DINO v2 \cite{DINOV2} with \cite{DINO_Reg} & SAM \cite{SAM} & CLIP \cite{CLIP} \\ \hline
		NMI (\%) $\uparrow$ & \textbf{55.10} & 17.16 & 33.16 \\
		ARI (\%) $\uparrow$ & \textbf{73.32} & 55.26 & 72.90 \\ \hline
	\end{tabular}
	\caption{Performance comparison of MPAE with different supervised pretrained backbones on the PartImageNet Segmentation dataset in the setting of $K=50$. The number of MPAE encoder and decoder layers is fixed at 2.}
	\label{Table12}
\end{table}

We also implement MPAE on the PartImageNet Segmentation dataset using backbones pretrained in a fully supervised manner, including the encoder of Segment Anything (SAM) and CLIP. The results are shown in Table 12. The training of SAM focuses on object boundaries rather than semantics, while CLIP mainly aligns instance-level descriptions with global ViT features in the latent space. Neither of them can produce finer-grained part-level features compared to DINO v2. Therefore, MPAE with DINO v2 achieves better performance, even though it is pretrained without any manual labels.

\section*{Appendix B.7 Influence of Mini-group Size}

\begin{table}[H]
	\centering
	\begin{tabular}{m{3.2cm}<{\centering}|m{1.0cm}<{\centering}m{1.0cm}<{\centering}m{1.0cm}<{\centering}m{1.0cm}<{\centering}|m{1.0cm}<{\centering}m{1.0cm}<{\centering}m{1.0cm}<{\centering}m{1.0cm}<{\centering}}
		\hline
		Dataset & \multicolumn{4}{c|}{CelebA} & \multicolumn{4}{c}{PartImageNet-S}  \\  \cline{1-9} 
		Mini-group size & 4 & 8 & 16 & 32 & 16 & 32 & 64 & 128 \\ \hline
		NMI (\%) $\uparrow$ & 59.50 & \textbf{59.64} & 53.89 & 25.52 & 43.94 & 47.28 & \textbf{55.10} & 53.63 \\
		ARI (\%) $\uparrow$ & \textbf{41.78} & 41.72 & 35.06 & 10.34 & 59.71 & 66.75 & \textbf{73.52} & 72.63 \\ \hline
	\end{tabular}
	\caption{Performance comparison of MPAE with different mini-group size on CelebA $(K=8)$ and PartImageNet Segmentation $(K=50)$.}
	\label{Table13}
\end{table}

We report the performance of MPAE with different mini-group sizes on a single-category dataset (CelebA) and a multi-category dataset (PartImageNet Segmentation) in Table 13 in Appendix B.7. With a very large mini-group size, MPAE tends to identify rarely appearing regions as independent parts. When the mini-group size is set to 32 on the CelebA dataset, the model identifies sunglasses and hands as independent parts instead of decomposing the face region into the target number of parts. Since we calculate the metrics using facial landmarks on the CelebA dataset, this results in a significant degradation in NMI and ARI on CelebA. Therefore, we set the mini-group size to 8 for datasets containing only a single category. However, different categories in PartImageNet Segmentation consist of various parts. A mini-group size that is too small forces MPAE to focus only on highly similar regions shared across different categories. Consequently, we observe a performance degradation when the mini-group size is less than 32 on datasets with multiple categories. As a result, we set the mini-group size to 64 for datasets with multiple categories. 

\section*{Appendix B.8 Average Number of Discovered Parts per Image with/without $\mathcal{L}_{\rm s}$}

\begin{table}[H]
	\centering
	\begin{tabular}{m{4.2cm}<{\centering}|m{1.8cm}<{\centering}m{1.8cm}<{\centering}}
		\hline
		Model & with $\mathcal{L}_{\rm s}$ & without $\mathcal{L}_{\rm s}$ \\ \hline
		Average number of discovered parts per image & 9.27 & 3.94  \\ \hline
	\end{tabular}
	\caption{Average number of discovered parts per image on PartImageNet Segmentation ($K=50$) with/without $\mathcal{L}_{\rm s}$}
	\label{Table14}
\end{table}

We compute the average number of discovered foreground parts per image with/without the constraint of $\mathcal{L}_{\rm s}$ to further investigate its influence, as shown in Table 14 in Appendix B.8. Without $\mathcal{L}_{\rm s}$, the average number of discovered foreground parts per image is only 3.94, indicating that each object is parsed into one or several coarse parts. This is because the MPAE without $\mathcal{L}_{\rm s}$ ssigns the $K$ parts primarily based on instance-level similarity rather than exploring the shared parts. $\mathcal{L}_{\rm s}$ encourages each part descriptor to respond only to regions with high semantic similarity on the feature map$\bm{F}$. As a result, the MPAE with $\mathcal{L}_{\rm s}$ can better discover the shared parts across multiple categorie, and the average number of the discovered foreground parts per image increases to 9.27.

\section*{Appendix C.1 Discovered Parts across Multiple Categories}

\begin{figure}[H]
	\centering
	\includegraphics[width=\linewidth]{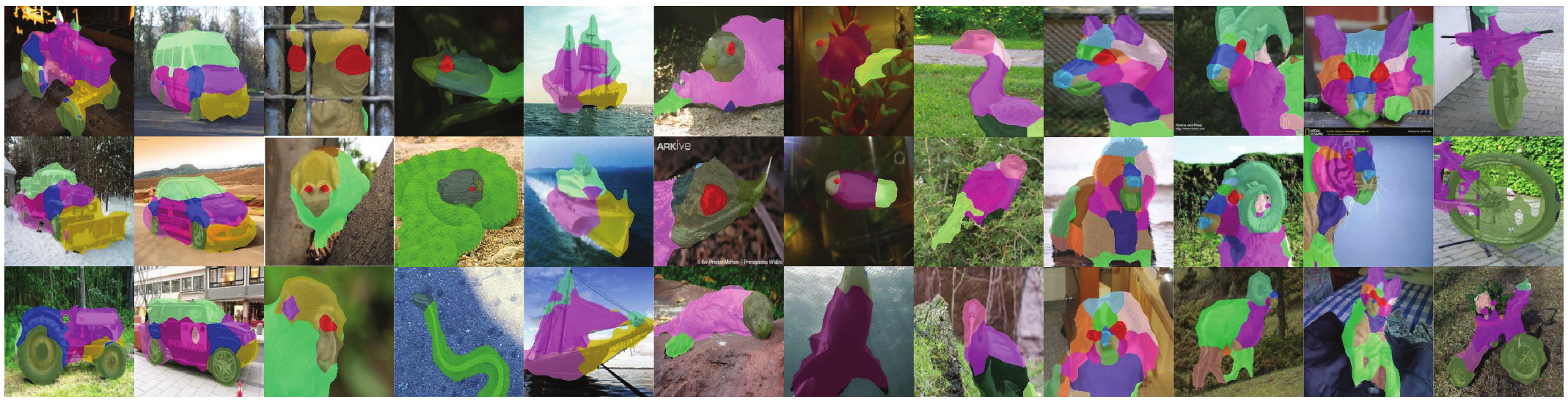}
	\caption{Parts unsupervisedly discovered by MPAE across multiple categories on PartImageNet Segmentation ($K=50$). The same color indicates that these discovered parts share similar semantics, even if they belong to different categories.}
	\label{fig7}
\end{figure}

\section*{Appendix C.2 Visualized Attention Maps in the Trainable ViT}

In Fig. 8, we present some pixel-level masks of the discovered parts predicted by MPAE, along with their corresponding attention maps from the trainable ViT used for descriptor extraction. The image restoration process performs implicit clustering, encouraging the features from the same part region to be similar. As a result, the region of each part on $\bm{S}$ is filled with the same part descriptor. By using these part descriptors to restore the masked patches, the ViT successfully learns to extract features from the corresponding part regions as part descriptors through the attention mechanism, as shown in Fig. 8 in Appendix C.2. This explains why the learned results can be robustly generalized to test images.   

\begin{figure}[H]
	\centering
	\includegraphics[width=\linewidth]{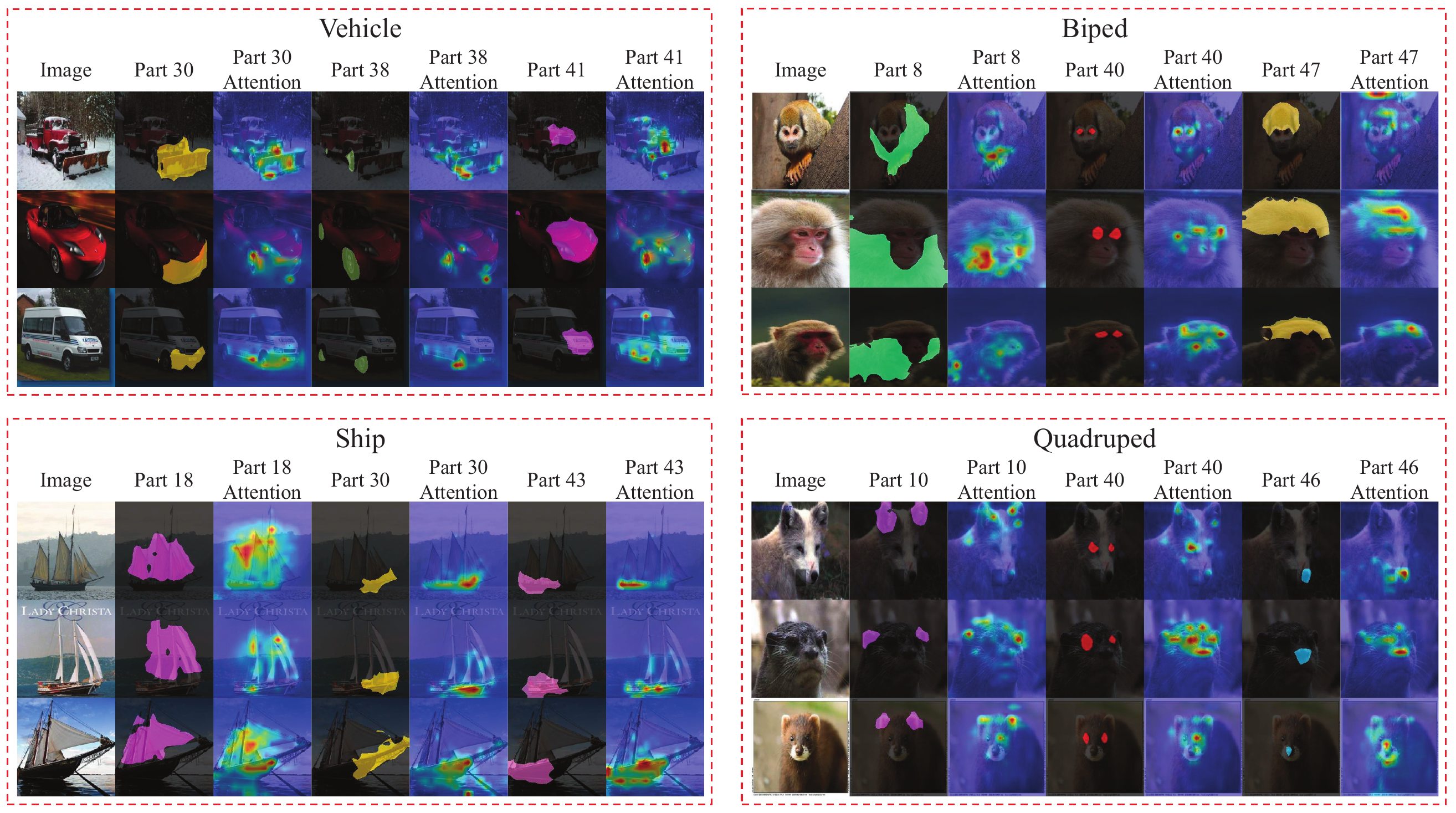}
	\caption{Pixel-level masks of discovered parts predicted by MPAE and their corresponding attention maps on PartImageNet Segmentation dataset ($K=50$).}
	\label{fig8}
\end{figure}

\section*{Appendix C.3 More Visualized Part Discovery Results}

\begin{figure}[H]
	\centering
	\includegraphics[width=\linewidth]{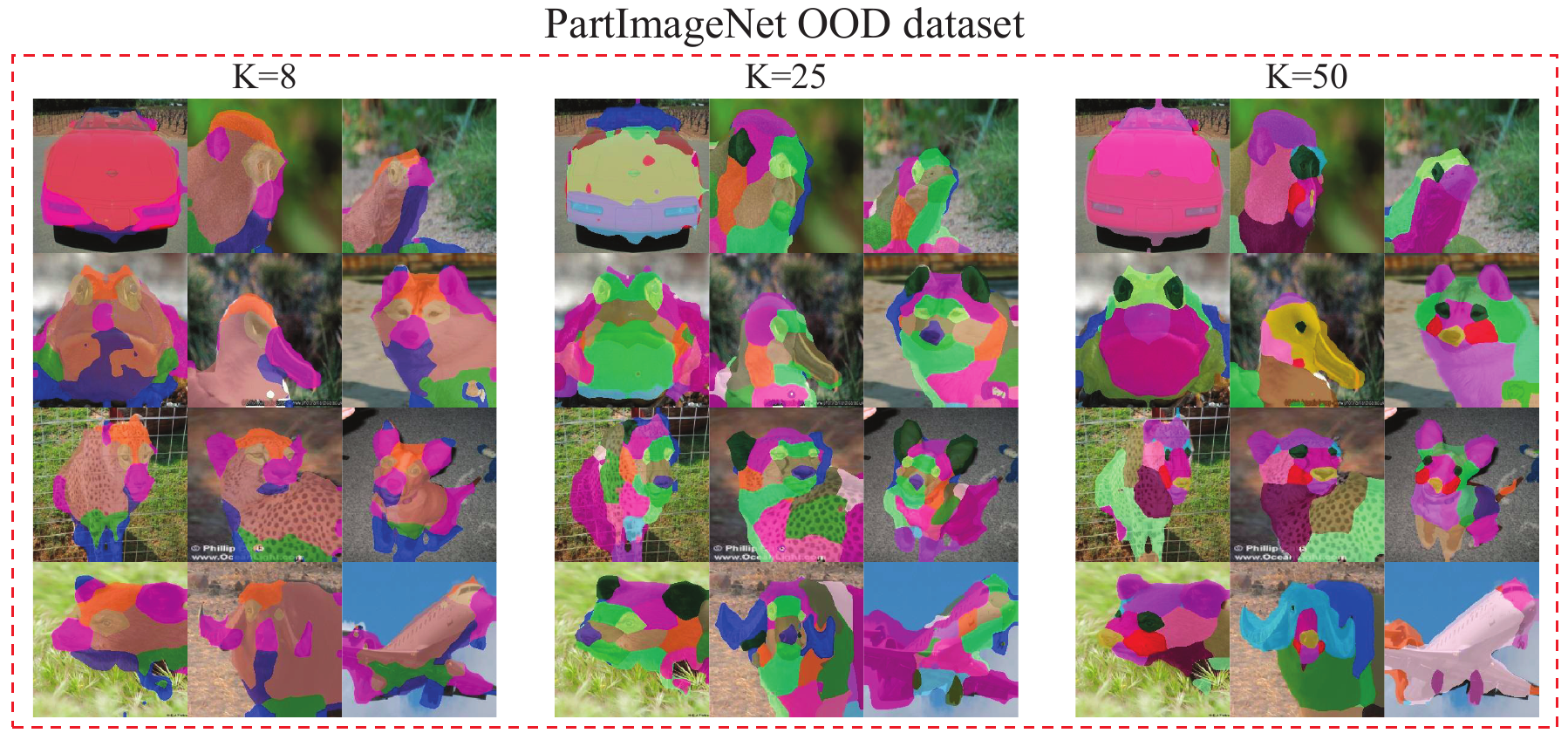}
	\caption{Examples of unsupervised part discovery results on PartImageNet OOD dataset predicted by MPAE in the setting of $K=8, 25, 50$.}
	\label{fig9}
\end{figure}

\begin{figure}[H]
	\centering
	\includegraphics[width=\linewidth]{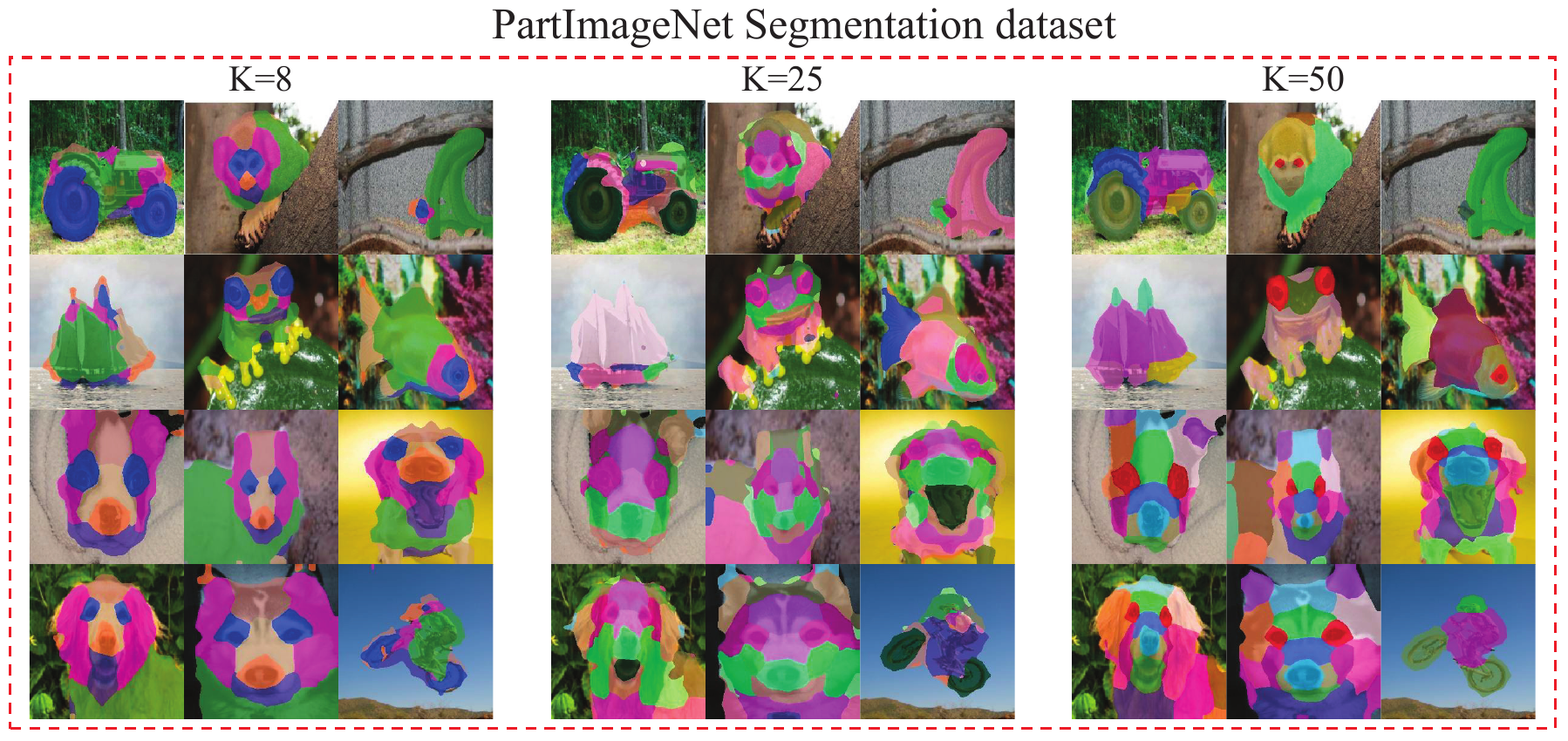}
	\caption{Examples of unsupervised part discovery results on PartImageNet Segmentation dataset predicted by MPAE in the setting of $K=8, 25, 50$.}
	\label{fig10}
\end{figure}

\begin{figure}[H]
	\centering
	\includegraphics[width=\linewidth]{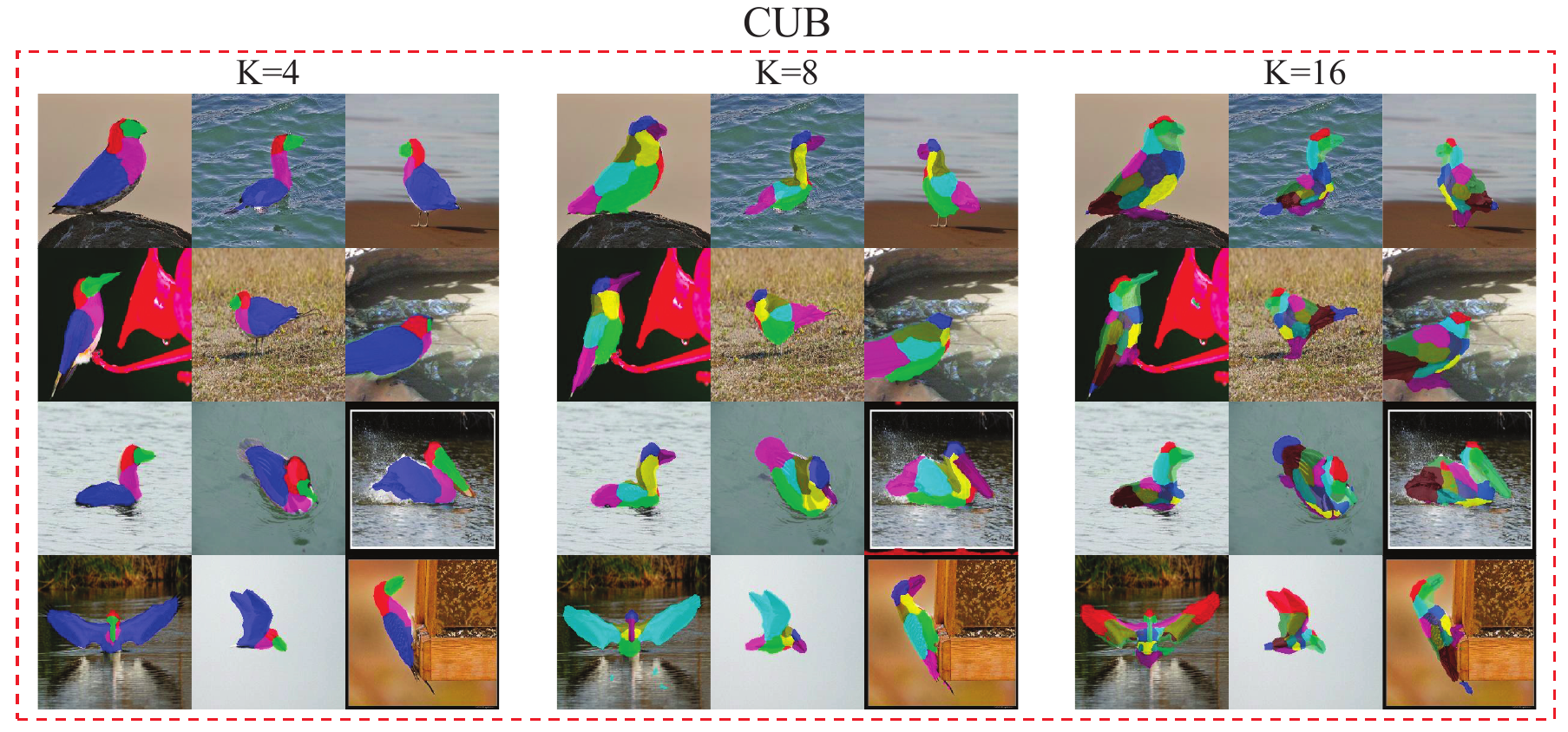}
	\caption{Examples of unsupervised part discovery results on CUB dataset predicted by MPAE in the setting of $K=4, 8, 16$.}
	\label{fig11}
\end{figure}

\begin{figure}[H]
	\centering
	\includegraphics[width=\linewidth]{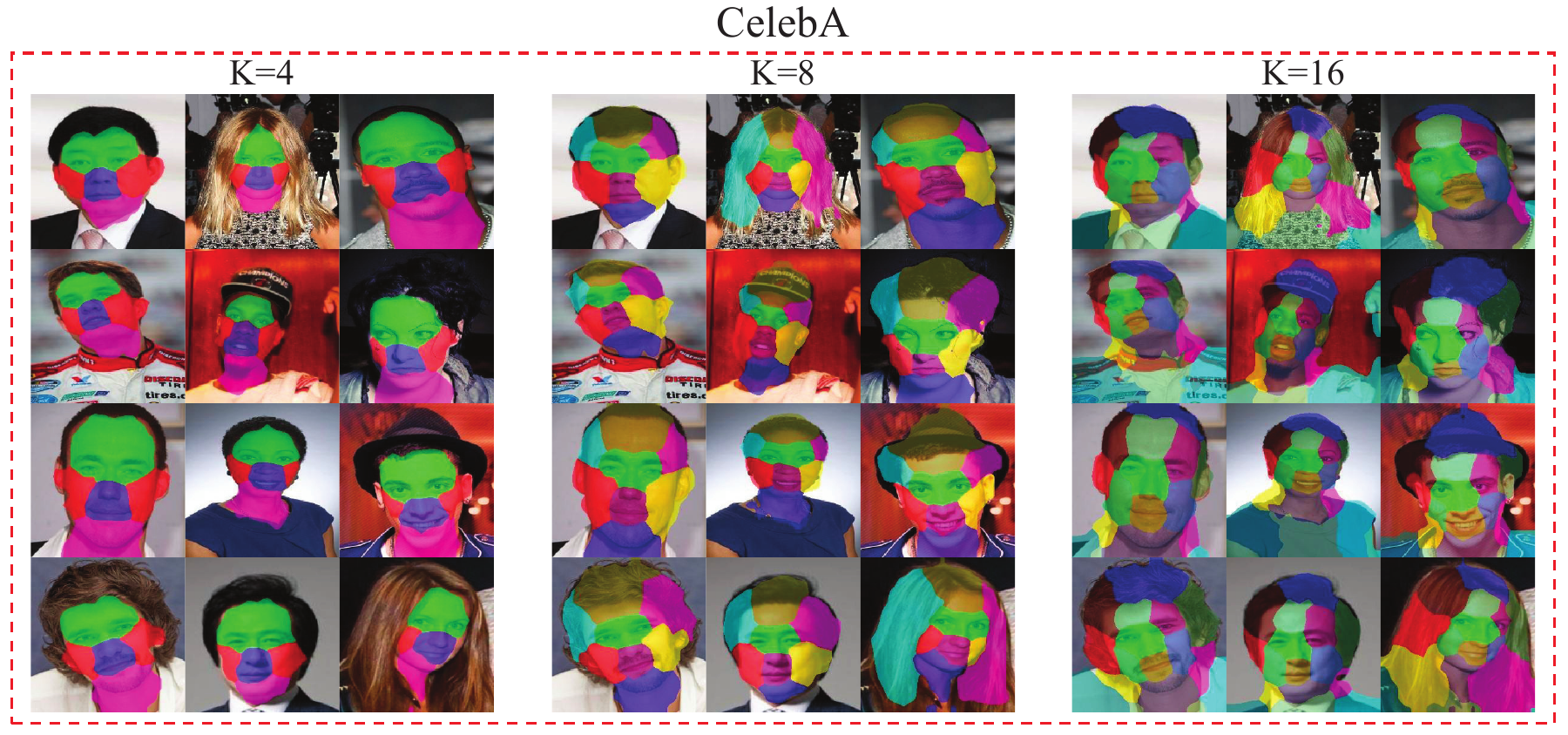}
	\caption{Examples of unsupervised part discovery results on CelebA dataset predicted by MPAE in the setting of $K=4, 8, 16$.}
	\label{fig12}
\end{figure}


\end{document}